\documentclass[lettersize,journal]{IEEEtran}
\usepackage{amsmath}
\usepackage{amsfonts}
\usepackage{algorithmic}
\usepackage{algorithm}
\usepackage{array}

\usepackage{textcomp}
\usepackage{stfloats}
\usepackage{url}
\usepackage{verbatim}
\usepackage{graphicx}
\usepackage{cite}
\usepackage{todonotes}
\usepackage{caption}
\usepackage{subcaption}
\usepackage{ltl}
\usepackage{hyperref}
\usepackage{xcolor}

\DeclareRobustCommand{\Ev}{\LTLonce}
\DeclareRobustCommand{\Gl}{\LTLpastglobally}

\DeclareRobustCommand{\Si}{\LTLsince}
\newcommand{\reals}{\mathbb{Re}}
\newcommand{\signal}{\mathbf{x}}
\newcommand{\ourmethod}{FERNN}
\newcommand{\edit}[1]{{\color{black} #1}}  
\newcommand{\fin}[1]{{\color{black} #1}}

\begin{document}


            


            

       
            

\title{Differentiable Inference of Temporal Logic Formulas}

\author{
Nicole Fronda and Houssam Abbas\\
Department of Electrical Engineering and Computer Science,
Oregon State University, Corvallis, OR\\
\{frondan, houssam.abbas\}@oregonstate.edu
\thanks{This work was partially funded by NSF Grant 1925652.}
}

\markboth{Journal of \LaTeX\ Class Files,~Vol.~14, No.~8, August~2021}%
{Shell \MakeLowercase{\textit{et al.}}: A Sample Article Using IEEEtran.cls for IEEE Journals}

\IEEEpubid{\begin{minipage}[t]{\textwidth}\ \\[8pt]
        \centering\normalsize{\fin{Manuscript received April 07, 2022; revised June 11, 2022; accepted July 05, 2022. This article was presented at the International Conference on Embedded Software (EMSOFT) 2022 and appeared as part of the ESWEEK-TCAD special issue.}}
\end{minipage}} 

\maketitle
\begin{abstract}
\fin{We} demonstrate \edit{the first} Recurrent Neural Network architecture for learning Signal Temporal Logic formulas, 
\edit{and present the first systematic comparison of formula inference methods.}
Legacy systems embed much expert knowledge which is not explicitly formalized. 
There is great interest in learning formal specifications that characterize the ideal behavior of such systems -- that is, formulas in temporal logic that are satisfied by the system's output signals. 
\edit{Such specifications can be used to better understand the system's behavior and improve design of its next iteration.}
Previous \edit{inference} methods either assumed certain \fin{formula templates}, or did a heuristic enumeration of all possible templates.
This work proposes a neural network architecture that infers the formula structure via gradient descent, eliminating the need for imposing \fin{any specific templates}. 
It combines learning of formula structure and parameters in one optimization.
\edit{Through systematic comparison,} we demonstrate that this method achieves similar or better mis-classification rates (MCR) than enumerative and lattice methods.
We also observe that different formulas can achieve similar MCR, empirically demonstrating the under-determinism of the problem of temporal logic inference.
\end{abstract}


\section{Introduction}
\IEEEPARstart{L}{egacy} systems encode a great amount of expert knowledge accumulated over time by the system's designers. 
This knowledge is implicit as often as it is explicit: while good design practice calls for Design Requirements documents to be created and referred to throughout the design and verification cycles, these requirements are still, more often than not, in natural language (e.g., English). 
Further, they leave much that is unspecified, and which is thus specified implicitly \fin{in the} design choices.
These choices are guided by the designer's intuition and training, and are not necessarily documented. 

The extraction of such implicit specifications from legacy systems, and turning them into \emph{explicit, formal} specifications has many benefits. First, it makes explicit the properties of a `good' design, as understood by the expert designers. This helps design future versions of the system.
These specifications can be used in model-checking and control synthesis algorithms, thus automatically improving the quality of future design versions. 
Finally, they can serve as formal contracts between interacting systems. Namely, this system guarantees that its outputs satisfy these extracted properties, which makes it possible to use it as a black-box when building a larger system in an Assume-Guarantee fashion \cite{sharf20AssumeGuarantee}.

In this paper, the formalism for expressing specifications is past-time Signal Temporal Logic (ptSTL).
Intuitively, ptSTL extends Boolean logic by introducing temporal modalities, like Historically $\Gl$ (the past counterpart of Always) and Once $\Ev$ (the past counterpart of Eventually), which allow us to speak of behavior over time. 

The main difficulty in learning a ptSTL specification from positive and negative examples of system behavior is the learning of the formula structure: e.g., 
What operators does it include? And in what order? Should it use a disjunction or conjunction? Etc.
This is a combinatorial search problem. 
We refer to this as the \emph{structure} of the formula. 
Thus $\Ev \Gl p$ and $\Ev \Gl q$ have the same structure, but $\Ev \Gl p$ and $\Gl \Ev p$ do not. 
State-of-the-art work either enumerates (almost) all possible structures and tries each one, or it endows the search space with some order (like a lattice).
Enumerative methods face a possibility of combinatorial runtime explosion, while lattice methods must confine themselves to a fragment of the logic.
In this paper, we develop a neural network-based approach to solving the structure learning problem without restrictions on the logic and without enumeration. 

The computation graph of a Recurrent Neural Network (RNN) is a natural representation of 
ptSTL robust semantics \cite{FainekosP09tcs,AbbasFSIG13tecs}.  This representation has been observed before \cite{LeungArechigaEtAl2020}.
We extend this representation to include choice over temporal and boolean operators, and let the network learn continuous weights between the various choices.
These continuous weights 
can be thought of as inducing soft choices between ptSTL's operators.
We combine this with a weight quantization procedure, which is incorporated in the training stage \cite{rastegari_XNOR} to ultimately learn binary weights. 
The final trained network thus implements the (robust) semantics of a ptSTL formula that can be simply read from the network. 
This contrasts with other works which learn a classifier or regressor, then try to extract an `explanation' from it \cite{Vardi21Neurosymbolic}.
The RNN also learns the atoms in the same training stage.
Thus a combinatorial search problem, structure inference, is solved with stochastic gradient descent.

In existing methods, the \fin{time} intervals decorating the temporal operators of the logic are learned \emph{after} a structure (also called a template) is chosen. 
Our proposed RNN-based method can also learn these intervals simultaneously with the structure learning, and we give examples. 
Because our focus is on solving the harder problem of structure learning, we focus our systematic analysis to infinite horizon formulas -- i.e., ones where the temporal operators have the unbounded intervals $[0,\infty)$.
In fact, our method can learn temporal constraints that are not continuous intervals, e.g., of the form $[0,1] \cup [5,7]$, and is the first method with this capability. Such disjoint intervals are a compact and more optimization-friendly way of representing conjunctions and disjunctions of constraints; for instance $\Ev_{[0,1]\cup [5,7]} = \Ev_{[0,1]}\lor \Ev_{[5,7]}$. This allows us to then perform the first systematic comparison between TL inference methods, including our proposed method. 
Our experiments use four datasets and several configurations.

One way to think of the Temporal Logic (TL) inference problem is as a classification problem in which we seek to learn a classifier with a special structure: namely, the classifier is a ptSTL formula.
There can be many different formulas that achieve similar mis-classification rate (MCR) on a given dataset. 
We present empirical evidence of this phenomenon. 
In the case of ptSTL inference, this under-determinism is particularly problematic: TL inference attempts to give a meaningful explanation of why some behavior is considered good, and other behavior is considered bad.
The idea is that the learned formula tells us something `real'  about the data.
However, if many different formulas yield the same MCR, and so are equally good explanations, then it is less reasonable to think of them as explanations at all. 
\edit{Thus, we also present qualitative examinations of formulas as part of our comparison.}


The paper is organized as follows:
related works are reviewed in Section \ref{sec:related}, and technical preliminaries on temporal logic and RNNs are provided in Section \ref{sec:background}. 
The computation of robust semantics by an RNN is covered in Section \ref{sec:tl as rnn cells}.
Section \ref{sec:choice and learning} describes our approach to training with quantized weights such that the resulting trained network immediately yields a formula that classifies the dataset. 
We call such a network built under this approach a \textit{Formula Extractable RNN (\ourmethod)}.
The experiments in Section \ref{sec:experiments} perform an extensive comparison of \ourmethod~with the enumerative and lattice methods, which are representative of the existing approaches to (S)TL inference. Section \ref{sec:discussion} presents further discussion of the validity of our approach given experimental results. 
Section \ref{sec:conclusion} concludes the paper.


\section{Related Works}
\label{sec:related}
The main challenge in TL inference is inference of the formula structure - that is, the choice and ordering of operators - rather than inference of their (continuous-valued) parameters. Here, `parameters' refers to the time bounds on the temporal operator and the thresholds in the atomic propositions. 

Most existing methods learn Signal Temporal Logic (STL) formulas~\cite{belta_tl_inference,mohammadinejad_interpretable_2019,dang20inference}, while the work in \cite{topcu21maxsat} explicitly targets LTL learning.
STL inference methods include the enumerative method of \cite{mohammadinejad_interpretable_2019}, which enumerates almost all formula structures, and the lattice method of \cite{belta_tl_inference} and follow-on works, which is restricted to a fragment of STL and searches over it by leveraging a lattice defined over the restricted search space.
Both 
approaches first propose a formula structure, then learn its parameters. If the resulting formula has too high an MCR, a new structure is proposed, and the cycle repeats.

Enumeration achieves good MCR in practice, but its runtime can increase dramatically as the inferred formula size is increased, as we will show. This is inherent to the method, which from a formula of a given size $L$ tries all ways of modifying it to get a formula of size $L$+1.\footnote{Formula size will be defined later, and we will see that different authors use different definitions.} 
The lattice method is restricted to reactive STL, which consists of formulas of the form $\Ev_{[\tau,\tau')}(\varphi_c \implies \varphi_e)$, where the cause and effect formulas $\varphi_c$ and $\varphi_e$ are only allowed to be conjunctions and disjunctions of delayed linear predicates, $\Ev_I \ell$ and $\Gl_I \ell$. 
Reactive STL has a partial order that is leveraged by the learning algorithm.
Moreover, both the enumerative and lattice methods work only with monotone formulas. 

Our proposed method \ourmethod~differs in that it combines inference for structure and atom thresholds, potentially avoiding the sub-optimality that arises from separating these two steps. 
Our method also does not require monotony with respect to the thresholds.
It handles the full LTL grammar, unlike the order-based lattice method.
Most notably, it leverages differentiable optimization to explore the combinatorial structure space, and represents up to $2^N$ formula structures in a compact RNN architecture with $N$ \edit{branching choices}.

The \textsf{samples2LTL} tool \cite{topcu21maxsat} uses MaxSAT to infer an LTL$_f$ formula (LTL for time-bounded signals) from a set of positive and negative traces. Unlike our method, it cannot learn formula size, or intervals on the temporal operators. Similar in spirit is the approach of \cite{majumdar20optsat} which uses OptSAT to learn a formula interactively with a human user. The technique in \cite{gol21repair} uses a pre-specified set of parametric formula templates from which to learn a root-cause for a system's failure.
Finally, the algorithm in \cite{dang20inference} takes a \textit{given} formula structure and computes the \emph{set} of parameters that achieve a given false positive and false negative rates, where possible.

\section{Background}
\label{sec:background}
\subsection{Temporal Logic Preliminaries}
\label{sec:tl prelims}
\subsubsection{Definition, Syntax, and Semantics}


Signal Temporal Logic (STL) \cite{maler_stl} is used for specification of desired system behaviors over time. This work uses ptSTL, a variant that specifies \textit{past} behaviors such as ``The robot \emph{was always} within 100 meters of its base station" or ``\emph{At one point in the last 5 minutes}, the robot returned to charge". 
\edit{We use ptSTL, rather than the more familiar future-time STL, to be consistent with the convention for RNNs to compute their outputs based on the previous outputs and inputs - not future ones. However, this paper's techniques are equally applicable to future-time STL; the choice of ptSTL is made for simplicity of presentation.}

The syntax of a ptSTL formula is defined recursively as:

\begin{equation*}
\phi := \top~|~\mu ~|~  \neg \phi  ~|~  \phi \lor \psi ~|~  \phi \Si_{I} \psi
\end{equation*}
where $\top$ is the constant True and $\mu$ is an atomic proposition from a set of atomic propositions $AP$. 
In operator $\Si_I$, $I$ is a subset of $\reals$, and for convenience we define $I_m := \inf I$ and $I_M := \sup I$.
The formula $\phi \Si_{I} \psi$ is read as ``$\psi$ held true at some point between $I_m$ and $I_M$ time units ago, and $\phi$ has held \emph{Since} then".  
Using the Since operator, two other temporal operators $\Ev$ and $\Gl$ are derived as:

\begin{equation*}
\begin{split}
    \Ev_{I} \phi & = \top \Si_{I} \phi \\
    \Gl_{I} \phi & = \neg \Ev_{I} \neg \phi \\
\end{split}
\end{equation*}

The operator $\Ev_{I}$ is the Once operator, and $\Gl_{I}$ is the Historically operator. 


We now define the semantics for ptSTL.  
It is a logic interpreted over signals $\signal:\reals \rightarrow \reals^d$.
For every $\mu$ in $AP$ we associate vector $W_\mu$ and scalar $b_\mu$.

\begin{eqnarray} \label{semantics atom}
    (\signal,t) &\models& \mu \text{ iff } W_\mu^T\signal[t]+b_\mu \geq 0
    \\
    (\signal,t) &\models& \neg \phi \text{ iff } (\signal,t) \not\models \phi
    \\
    (\signal,t) &\models& \phi_1 \lor \phi_2 \text{ iff } (\signal,t) \models \phi_1 \text{ or }  (\signal,t)\models \phi_2
    \\
    (\signal,t) &\models& \phi \Si_I \psi \text{ iff } \exists t' \in t-I \text{ s.t. } (\signal,t')\models \psi \nonumber
    \\
    &&\text{ and }  \forall t'' \in [t',t)~(\signal,t'') \models \phi
\end{eqnarray}
For instance we can now derive
\begin{eqnarray} \label{semantics ev}
    (\signal,t) &\models& \Ev_I  \text{ iff } \exists t' \in t-I \text{ s.t. } (\signal,t') \models \phi
    \\
    (\signal,t) &\models& \Gl_I  \text{ iff } \forall t' \in t-I \text{ s.t. } (\signal,t') \models \phi
\end{eqnarray}
In words, $\Ev_{[a,b]}\phi$ says that $\phi$ held (at least) once between $a$ and $b$ time units in the past, while $\Gl_{[a,b]}\phi$ says that $\phi$ held continuously between times $a$ and $b$.
Note that atomic propositions are given by linear functions of the state.

\subsubsection{Robustness}
A quantitative measure of how well a ptSTL formula is satisfied by a signal is given by the \textit{robustness}. 
The robustness is a conservative estimate of the distance to violation of the formula by the signal at time $t$. 
Robustness of signal $\signal$ to $\phi$ at time $t$, denoted by $\rho_{\phi}(\signal, t)$ is defined as~\cite{FainekosP09tcs}. 
\begin{equation} \label{rob semantics atom}
    \rho_{\mu}(\signal, t)  = W_\mu^T\signal[t]+b_\mu
\end{equation}
\begin{equation} \label{rob semantics negation}
    \rho_{\neg \phi}(\signal, t) = - \rho_{\phi}(\signal, t)
\end{equation}
\begin{equation} \label{rob semantics or}
    \rho_{\phi \lor \psi}(\signal, t) = \max(\rho_{\phi}(\signal, t), \rho_{\psi}(\signal, t))
\end{equation}
\begin{equation} \label{rob semantics since}
    \rho_{\phi \Si_I \psi}(\signal, t) = \sup_{t' \in t-I}\left(
    \min(\rho_{\psi}(\signal, t'),
    \inf_{t'' \in [t',t)}(\rho_{\phi}(\signal, t'')) \right)
\end{equation}
Thus we derive that $\rho_{\Ev_I \phi}(\signal, t) = \sup_{t' \in t-I}\rho_{\phi}(\signal, t')$ and $\rho_{\Gl_I \phi}(\signal, t) = \inf_{t' \in t-I}\rho_{\phi}(\signal, t')$.

We will give examples of our method learning bounded time intervals $I$.
But to focus on the structure inference problem, and unless otherwise stated, by default in this paper we consider \textit{time-unbounded} formulas. That is, we set $I=[0,\infty)$, and drop it from the notation.

\subsection{Recurrent Neural Networks}
\begin{figure}[!t]
\centering
\includegraphics[width=\linewidth]{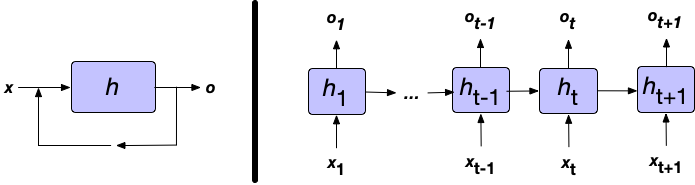}
\caption{Example of an RNN cell in compact form (left) and unfolded (right). The input $x_t$ is used with hidden state $h_t$ to produce output $o_t$ for every $t$ element of the input sequence.}
\label{fig:rnn_block_example}
\end{figure}

A Recurrent Neural Network (RNN) is a type of artificial neural network that uses its hidden states to retain memory of past inputs for use with current inputs to generate its output \cite{rnn_hochreiter}. These hidden states, or memory units, enable processing of variable length sequential data. Figure \ref{fig:rnn_block_example} illustrates a simple example of a single RNN cell which takes as input a sequence of length $T$, $x = [x_1, \ldots, x_{t-1}, x_t, x_T]$ and for each element of the sequence produces an output $o_t$ using a hidden value $h_t$. The general functions for computing $h_t$ and $o_t$ are:

\begin{eqnarray*}
    h_t &=& \sigma_h (W_h x_t + U_h o_{t-1} + b_h) \\
    o_t &=& \sigma_o (W_o h_t + b_o) \nonumber
\end{eqnarray*}
where the $W$'s and $U$'s are row vectors, $b$'s are scalars, and the $\sigma$ are activation functions like sigmoids. 

\section{Learning Formulas with \ourmethod}
\label{sec:learning formulas with rnns}
\label{sec:rnn cells}
 \ourmethod~is an RNN architected to compute the robustness of any ptSTL formula
(Section \ref{sec:tl as rnn cells}). 
We can, in fact, have a single \ourmethod~compute the robust semantics of any given number of formulas, by quantizing the values of the learnable weights in the network (Section \ref{sec:choice and learning}).
Assuming for now that our dataset consists of signals and their robustness values relative to some unknown formula, we train the network (that is, optimize weights and biases) to learn this robustness function.
We can then read the learned formula by simple inspection of the final trained network. 
In practice, robustness values are not available -- after all, the formula is not available and we are trying to infer it. 
Thus our datasets actually consist of signals and their binary labels: +1 for positive examples, and -1 for negative examples.
We will see that even this discrete labeling is enough for our method to perform well.

\subsection{Temporal Logic Operators as RNN Cells}
\label{sec:tl as rnn cells}
\begin{figure*}[ht]
     \begin{tabular}{cccc}
        {\includegraphics[width=.25\linewidth]{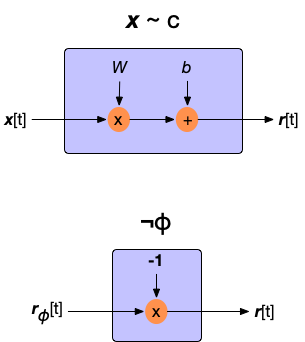}}
        & 
        {\includegraphics[width=.23\linewidth]{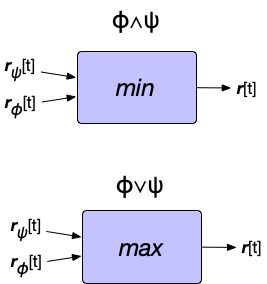}}
        &
        {\includegraphics[width=.24\linewidth]{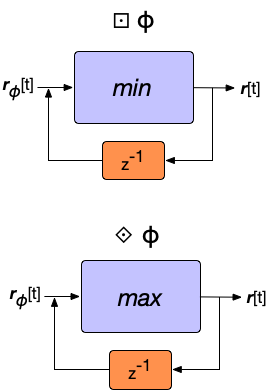}}
        & 
        {\includegraphics[width=.25\linewidth]{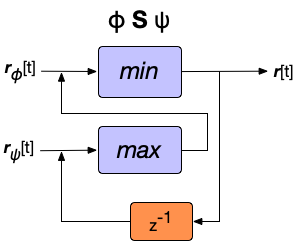}}
    \end{tabular}
    \caption{Illustration of STL operators as RNN cells as defined in Section \ref{sec:tl as rnn cells}}.
    \label{fig:rnn_stl_cells}
\end{figure*}

In this section, we show how the robust semantics of a given formula $\phi$ can be computed by a \ourmethod.
This forms the basis for the next section, which shows how the \ourmethod's weights can be trained to learn a formula classifier from a given dataset. 
The cells of a \ourmethod~are illustrated in Fig. \ref{fig:rnn_stl_cells}. 
A model comprised of these cells takes as input a signal $\signal$ and outputs robustness $\rho_\phi(\signal,\cdot)$.
The hidden neurons take robustness signals as inputs (computed by previous layers) and output other robustness signals.
In this section we fix the signal $\signal$.
We will write $\phi[t]$ instead of $\rho_{\phi}(\textbf{x}, t)$ for the robustness signal input to a hidden neuron, and $r_\phi[t]$ for the output robustness signal from a hidden neuron.
Thus in this notation, the \ourmethod~computing $\rho_{\Ev p}$ has a hidden neuron that receives $p[t]$ and outputs $r_{\Ev p}[t]$.

\paragraph{Atomic Proposition} The robustness $\mu[t]$ is easily computed by a single neuron with bias $b_\mu$ and input weight vector $W_\mu$ and identity activation. 

\paragraph{Negation} This is trivially implemented by a neuron with one input $\phi[t]$, (scalar) weight $W=-1$, 0 bias, and identity activation.

The following boolean and temporal operators fit the general equation for output of an RNN cell with a weights vector of all 1s and bias of 0.

\paragraph{Conjunction and disjunction} 
These are implemented with neurons that apply max/min activations to two inputs:
\begin{eqnarray} \label{and_or_to_neuron}
    r_{\phi \land \psi}[t] &=& \min(\phi[t], \psi[t])
    \\
    r_{\phi \lor \psi}[t] &=& \max(\phi[t], \psi[t]).
\end{eqnarray}

For all the above neurons, no memory units are required.

For the temporal operators, we present the unbounded versions here for conciseness; Section \ref{sec:exp time bounds} of the Experiments will illustrate how we can also learn temporal intervals.
\paragraph{Once and Historically}
We compute $\Ev \phi$ and $\Gl \phi$ as recurrent cells with a single memory unit and functions with max/min activations given a single input.
\begin{equation} \label{alw_to_cell}
    r_{\Gl \phi}[t] = \max(\phi[t], r_{\Gl \phi}[t-1])
\end{equation}

\begin{equation} \label{ev_to_cell}
    r_{\Ev \phi}[t] = \min(\phi[t], r_{\Ev \phi}[t-1])
\end{equation}



\paragraph{Since}
Following the robust semantics given in Eq. (\ref{rob semantics since}) we represent $\phi \Si \psi$ as a recurrent cell with a single memory unit and function given two inputs:

\begin{equation} \label{since_to_cell}
    r_{\phi \Si \psi}[t] = \min(\phi[t], \max(r_{\phi \Si \psi}[t-1], \psi[t]))
\end{equation}

The derivation for Eq. \eqref{since_to_cell} from the semantics of the $\Si$ operator is provided in the Appendix.

A few important remarks are in order:
it is obvious that we can implement any formula's robustness as an RNN using these cells, so we are not constrained to a fragment of the logic.
In the other direction, given such an RNN, we can simply read off the corresponding formula -- no need for a subsequent `extraction' or `explanation' step.
The number of layers of the \ourmethod~directly measures the complexity of the formula (number of nesting levels); thus we have a convenient handle on the complexity/goodness-of-fit trade-off.

\subsection{Choice Blocks and Learning Formula Structure} \label{sec:choice and learning}
If the formula structure, and therefore the corresponding \ourmethod~architecture, is pre-fixed, it is trivial to learn the parameters $W_\mu$ and $b_\mu$ of the atoms.
\fin{However,} our goal is to learn the structure itself. 
We introduce new cells called \textit{choice blocks} that contain learnable weights \fin{enabling} \ourmethod~to choose which operators or atomic propositions belong in the formula.  \edit{In typical RNNs, weights are continuous and real-valued. To translate these weights to discrete choices, we apply a quantization within the choice blocks.} 

\begin{figure}[!t]
\centering
\includegraphics[width=0.6\linewidth]{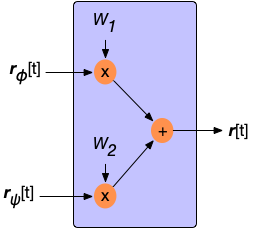}
\caption{Basic structure of a \ourmethod~choice block. Inputs $r_\phi$ and $r_\psi$ represent robustness of two possible formulas. \edit{Input $r_\phi$ is multiplied by weight $W_1$ and input $r_\psi$ is multiplied by weight $W_2$. A one-hot quantization is applied to weights $W_1$ and $W_2$ after learning, so exactly one of them is 1 and the other 0.} Thus the output $r$ equals exactly $r_\phi$ or $r_\psi$. The input with the non-zero quantized weight is ``chosen" for the formula.}
\label{fig: choice block}
\end{figure}
Figure \ref{fig: choice block} illustrates an example choice block.  It takes as input the robustness output signal of two \ourmethod~cells (though any number of inputs is possible) and yields a single robustness value from exactly one of the cells, essentially `choosing' which of the cells' information is passed forward. This selection is done by learning a weight $W_i$ for each choice block input. 
Exactly one weight will be either 1 or -1, and the rest will equal 0.  \edit{The input assigned the non-zero weight is `chosen' to be included in the formula. Now, how do we learn these discrete weights?  In fact, the learned real-valued weights are transformed into discrete, one-hot weights via quantization.}


The output of the choice block is then calculated as:
\begin{equation} \label{choice_block}
    r[t] = \sum_{n=1}^{N}\alpha B_i r_i[t] 
\end{equation}
where $B_i$ is the quantized value of $W_i$ and $\alpha$ is a scaling factor such that $W \approx \alpha B$. 
Thus, because all but one of the $B_i$ equal 0, only the input with the non-zero quantized weight is passed forward.

\paragraph{Learning one-hot weights} We modify the binary quantization method proposed in \cite{rastegari_XNOR} to implement this one-hot constraint. Let $W$ be a real-valued weight vector and $B$ be the corresponding one-hot quantized weight vector. 
To find an optimal value for $B$ and scalar $\alpha$ such that $W \approx \alpha B$ for some $\alpha > 0$, we solve the following optimization.

\begin{eqnarray}
    \label{eq: opt_bin_weights}
    J(B, \alpha) &=& ||W - \alpha B||^2 \\
    \alpha^*, B^* &=& \text{argmin}_{\alpha, B} ~J(B, \alpha) \nonumber
\end{eqnarray}

By expanding Eq. (\ref{eq: opt_bin_weights}) above we have
\begin{eqnarray}
    J(B, \alpha) &=& \alpha^2 B^{T}B - 2\alpha W^{T} B + W^{T} W \nonumber
\end{eqnarray}

Because B is one-hot, $B^{T}B = 1$. Additionally, $W^{T}W$ is a known constant. Setting $W^{T}W = c$, we now have 
\begin{eqnarray}
    J(B, \alpha) &=& \alpha^2 - 2\alpha W^{T} B + c \nonumber
\end{eqnarray}
Because $\alpha$ is positive, minimizing $J$ requires maximizing the second term. Without loss of generality, we can first find the optimal $B^*$ for a generic positive $\alpha$:

\begin{equation}
    \label{eq: optimal B}
    B^* = \text{argmax}_{\text{one-hot } B} ~ W^{T} B
\end{equation}
The optimizer is $B^*_j = sign(W_j)$ where $W_j = \max(W)$ and $B^*_{i\neq j} = 0$.
Then, to find the optimal $\alpha^*$, we take the derivative of $J$ with respect to $\alpha$ and set it to 0:

\begin{equation}
    \label{eq: optimal alpha}
    \alpha^* = W^{T} B^* = \max(W) 
\end{equation}

During training of the network, the real-valued weights are quantized using Eqs. \eqref{eq: optimal B},\eqref{eq: optimal alpha} before performing the forward pass.
The output of the forward pass is calculated using the quantized weights according to Eq. (\ref{choice_block}). 
During backpropagation, the gradients computed from this output are used to update the real-valued weights.

In general, quantizing the weights of a neural network decreases accuracy due to loss of information when approximating of the real-valued weights. However, in practice we found this difference in performance can be minimal -- an acceptable trade-off for extracting a readable, logical representation of the trained network. Table \ref{tab: quantization comparison} in the Appendix shows the effects of quantization in our method.

\begin{figure*}[ht]
     \begin{tabular}{ccc}
        \subcaptionbox{}
        {\includegraphics[width=.25\linewidth]{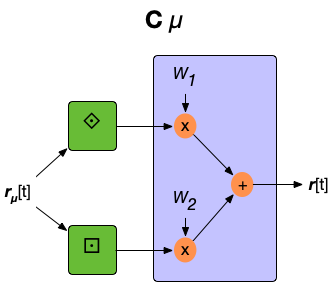}}
        & 
        \subcaptionbox{}
        {\includegraphics[width=.3\linewidth]{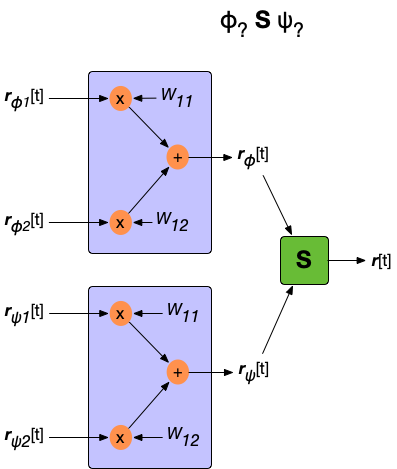}}
        &
        \subcaptionbox{}
        {\includegraphics[width=.36\linewidth]{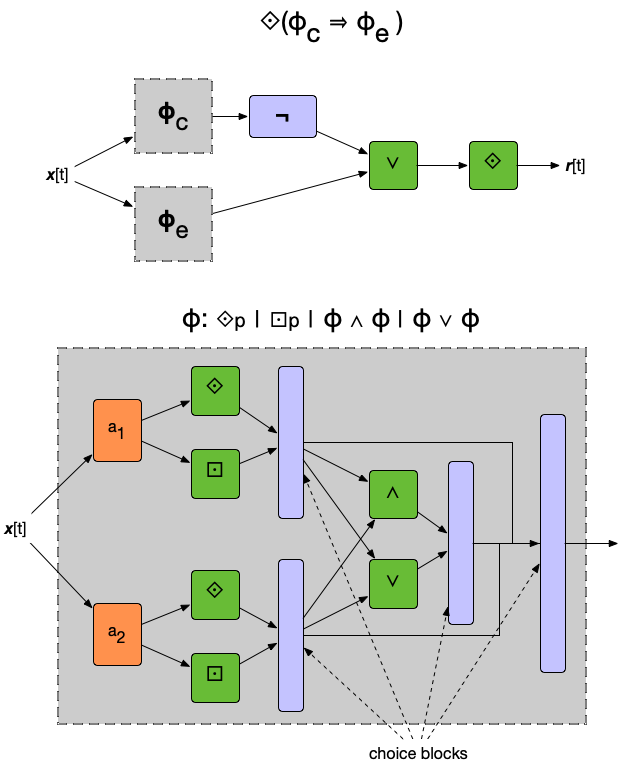}}
    \end{tabular}
    \caption{Examples of using choice blocks and \ourmethod~cells to construct networks that learn the structure and parameters of STL formulas. (a) learns a formula of the form $C \mu$ where $C$ is a choice between $\Ev$ and $\Gl$, (b) learns a formula of the form $\phi \Si \psi$ where each $\phi$ and $\psi$ have two possibilities, and (c) learns a formula given in reactive PSTL (defined in \cite{belta_tl_inference}).
    }
    \label{fig:rnn_examples}
\end{figure*}

Finally, the steps to learn and extract a formula as a binary classifier for a data set using \ourmethod~cells are as follows:
\begin{enumerate}
    \item \edit{Obtain a set of signals labelled with robustness values. Positively labeled signals are examples of desired behavior, and negatively labeled signals are examples of bad behavior. Note that typically, it is not always possible to obtain actual robustness values for the training signals since that assumes that we already know a ground truth formula (which defeats the purpose a learning one). Other quantitative measures that may approximate robustness can be used in practice (e.g., scores given by experts), but in the general absence of such values, binary labels can be used. With binary labels, +1 labeled signals are examples of desired behavior, and -1 labeled signals are examples of bad behavior.}
    \item Construct a network architecture with the \ourmethod~cells. This step can be very flexible: a network with $N$ choice blocks embeds $2^N$ possible formulas. Prior domain knowledge may be used to impose a structure for the final formula: e.g., if we expect that a formula in reactive PSTL \cite{belta_tl_inference} is called for, we can use the architecture in Fig.~\ref{fig:rnn_examples}(c). But choice blocks can also be used to construct a much more general structure space and let \ourmethod~learn the appropriate structure simultaneously with the atom parameters.
    \item Train the network by using quantized weights in the forward pass and updating the real-valued weights during backpropagation. Loss is calculated using the difference between predicted robustness and the data labels. If binary labels are used instead of true robustness values, a non-linear activation is applied to the final layer output to force positive robustness outputs near 1 and negative robustness outputs near -1.
    \item After training, recover the learned atomic propositions and the chosen (non-zero) inputs from each choice block to construct the formula. 
\end{enumerate}

\section{Experiments}
\label{sec:experiments}

This section describes application of our method to classification problems on different datasets. We compare our method's performance against two existing temporal logic inference methods: 1) an Enumerative method \cite{mohammadinejad_interpretable_2019} and a 2) Lattice based method \cite{belta_tl_inference}. 
In the Enumerative method, all possible formula structures up to a pre-specified formula length are enumerated and stored in a database.  For each formula structure in the database, different values for the parameters are tested until a target mis-classification rate (MCR) against a holdout dataset is met. In the Lattice method, possible formula structures for a reactive PSTL formula are organized as nodes in a graph. Pruning and growing of the graph refines the search for the best formula structure. For each node in the graph, different parameter values are tested until a satisfactory MCR is found.  We refer the reader to the respective papers of each method for further details.  We evaluate \ourmethod~against these methods in terms of MCR and wall clock runtime.  We also discuss the quality of the best formulas found by each of the methods based on domain knowledge of the datasets. Finally, we demonstrate how our method can also learn temporal intervals for the operators.

\edit{\textbf{Past vs Future STL:} We presented our method for learning past STL formulas, whereas the Enumerative and Lattice methods are formulated to learn future STL formulas.  To compare the methods, we feed signals in reverse chronological order to the \ourmethod~models to generate future STL formulas.}  

\subsection{Datasets}

\paragraph{Cruise Control of a Train (CCT)} We recreate the train cruise control experiment from \cite{mohammadinejad_interpretable_2019}. The cruising speed of a three-car train is set to $25m/s$ with oscillations $\pm2.5m/s$. Under normal conditions, the train maintains its cruising speed and applies brakes when the speed exceeds some upper limit. Anomalous conditions are simulated by disabling the train's brake system. We generate traces for the velocity signal $v$ of the train from 2000 simulations, half under normal conditions and half under anomalous conditions, which are given labels $+1$ and $-1$ respectively. Additionally, we generate continuous labels for each of the traces by calculating robustness at $t=0$ for each trace with respect to the formula $\textbf{G}(v <= 34.2579)$, which was the final learned STL formula in \cite{mohammadinejad_interpretable_2019}. 

\paragraph{Lyft Vehicle Trajectories} We attempt to learn a formula description for vehicle trajectories from the Lyft Prediction dataset \cite{lyft2020dataset} by classifying the true Lyft trajectories from artificial trajectories. 

The features of the traces provided by the Lyft dataset include the $x$ and $y$ coordinates of a vehicle's ground truth trajectory relative to its starting position on a semantic map. Two additional features were derived, $x_\textit{diff}$ and $y_\textit{diff}$, which capture the change in coordinate values between time steps. Artificial trajectories were generated by randomly sampling trajectories from the Lyft dataset and adding a fixed scalar uniformly sampled from range $(-20, 20)$ to the $x$ values and similarly for the $y$ values. The artificial trajectories are given label $-1$ and the true trajectories are given label $+1$.

\paragraph{Electrocardiogram (ECG)}
We use the processed ECG data from \cite{ecg_data} to train a classifier for normal heart rate. We use 131 traces of 10-second ECG signal fragments exhibiting normal sinus rhythm (label $1$) and 133 traces of equal length exhibiting atrial fibrillation (label $-1$). From the raw ECG signal, we also derive a feature for change in heart rate ($hr_\textit{diff}$) from the previous time step. Additionally, as with the CCT dataset, we generate continuous labels for each of the traces by calculating robustness at $t=0$ for each trace with respect to the formula $\textbf{G}(hr_\textit{diff} \leq 5.795)$, which was the final learned STL formula by the Enumerative method in our experiments.

\paragraph{Human Activities and Postural Transitions (HAPT)}
We use the smart phone data from \cite{ReyesOrtiz2016TransitionAwareHA} to train a classifier for dynamic activities (eg. walking) over static activities (eg. sitting, lying down). We use fifty dynamic and fifty static 10-second traces of standard deviation of acceleration signal in the x-direction collected from the smart phone triaxial accelerometer. Dynamic traces of this signal are given label $+1$ and static traces are given label $-1$.

\subsection{Experimental Setup} \label{sec: exp setup}
We trained five different \ourmethod~model architectures on each dataset to learn formulas of lengths 2 to 6.  \edit{That is, there is a model to learn formulas of length 2, another model that learns formulas of length 3, etc, and larger models incorporate the architectures of smaller models.}  We define length of a formula as the count of atomic propositions and operators to follow the convention used by the Enumerative method. In the Lattice method, formula length is defined as the number of atomic propositions in the formula. The formula $\phi = \Ev (x \geq c ~\land~ y \geq d)$ has length 4 according to the convention we follow, but length $2$ in the Lattice convention.  \edit{We choose a maximum formula length of 6 for practicality of our experiments, though a \ourmethod~may be designed to learn any arbitrary length formula.}

For classification tasks with $\pm 1$ labels, the final layer of each \ourmethod~model was passed into a tanh activation layer.  We used mean absolute error as our loss function and an Adam Optimizer with initial learning rate of 0.003.  Each dataset was split 80-20 for training and testing and normalized between 0 and 1 before passing to \ourmethod. We did not apply this normalization to the Enumerative and Lattice methods. 
Instead, we specified a search range for the parameters of the atomic propositions based on the feature values. 
We forced the Enumerative and Lattice methods to skip the search over time bounds and set each temporal operator to be time-unbounded, to focus on analyzing structure learning. 

Both the Enumerative and Lattice methods terminated their search when finding a formula with MCR $<0.2$. \ourmethod~stopped training early if loss did not decrease in the last 50 epochs, otherwise training ceased after 5000 epochs. 

The \ourmethod~models were implemented in Python.
For the Enumerative and Lattice methods, we used the MATLAB implementations provided by the authors. Code and data available at \url{https://github.com/nicaless/fernn_stl_inference}.

\subsection{Quantitative Comparison with Existing Methods}

\begin{figure*}[ht]
    \centering
     \begin{tabular}{cccc}
        {\includegraphics[width=.24\linewidth]{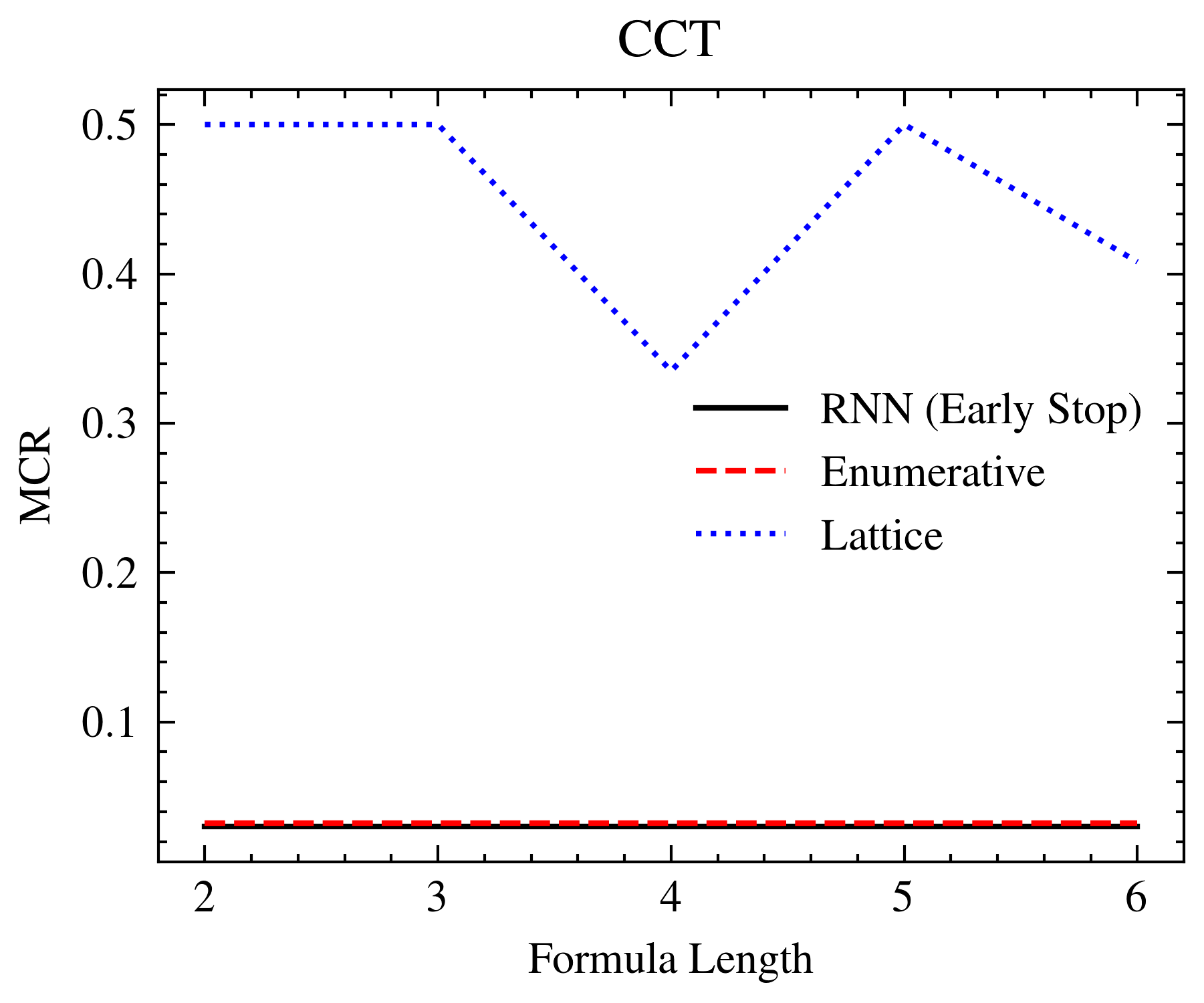}}
        & 
        {\includegraphics[width=.24\linewidth]{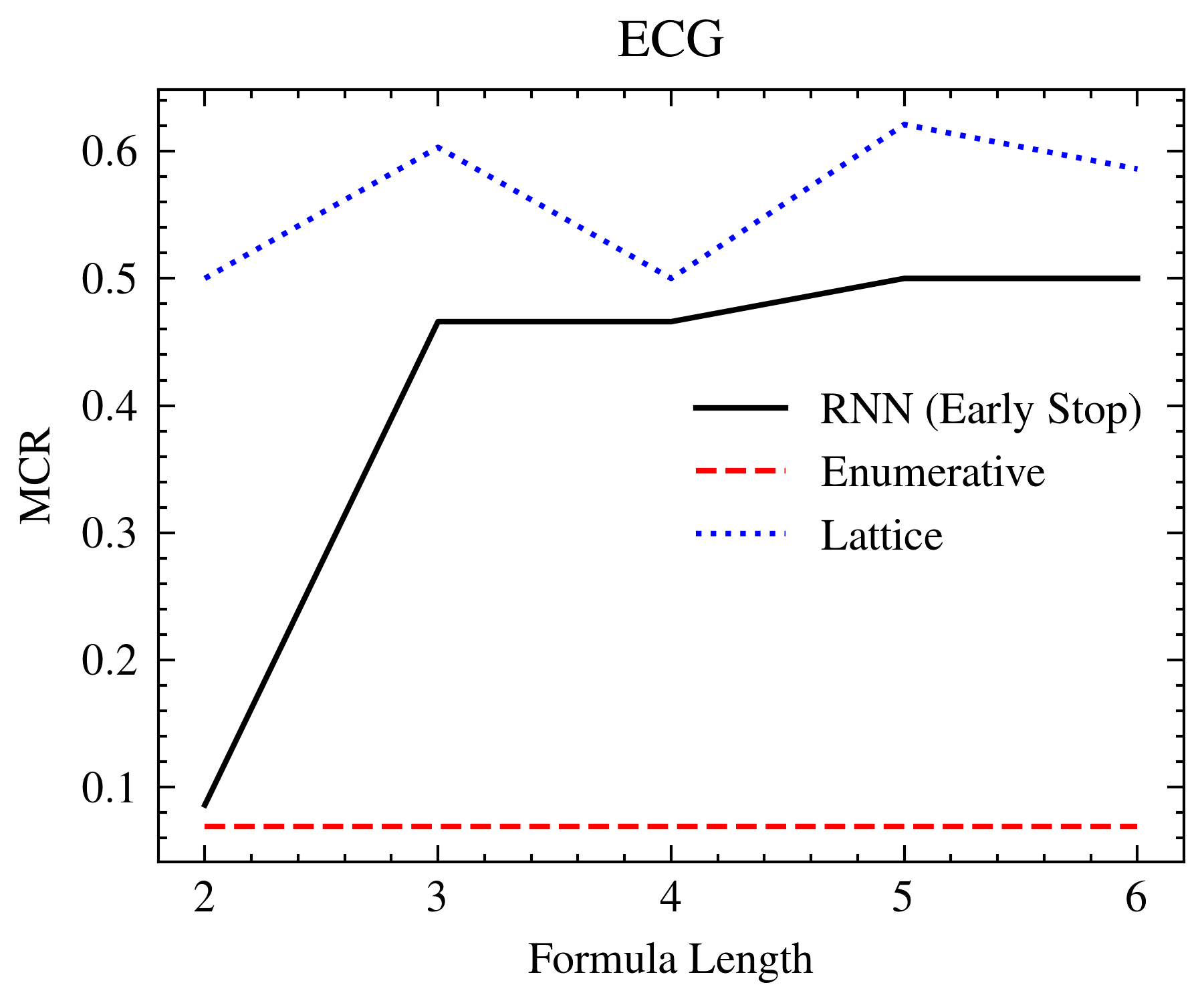}} 
        &
        {\includegraphics[width=.24\linewidth]{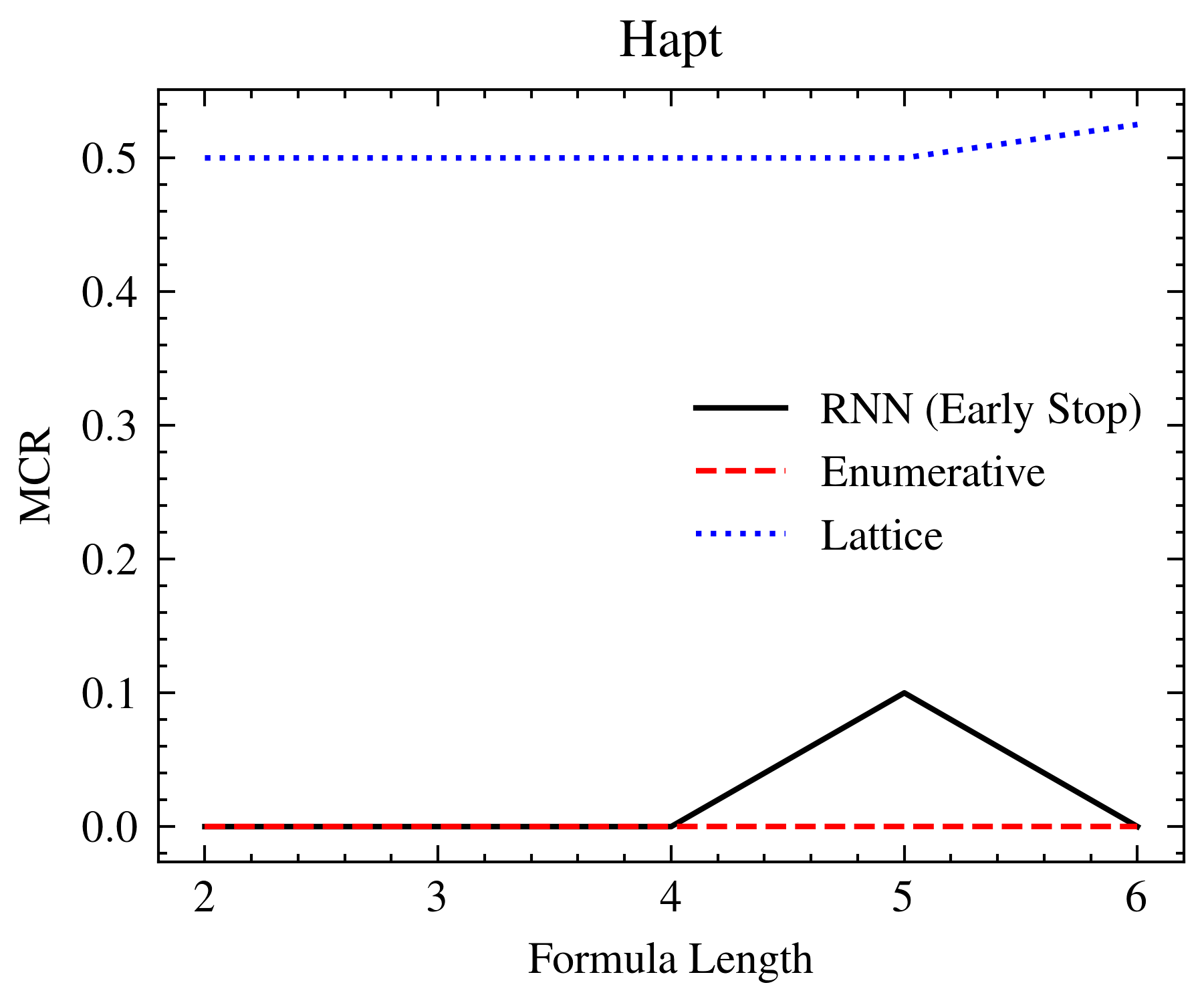}}
        & 
        {\includegraphics[width=.24\linewidth]{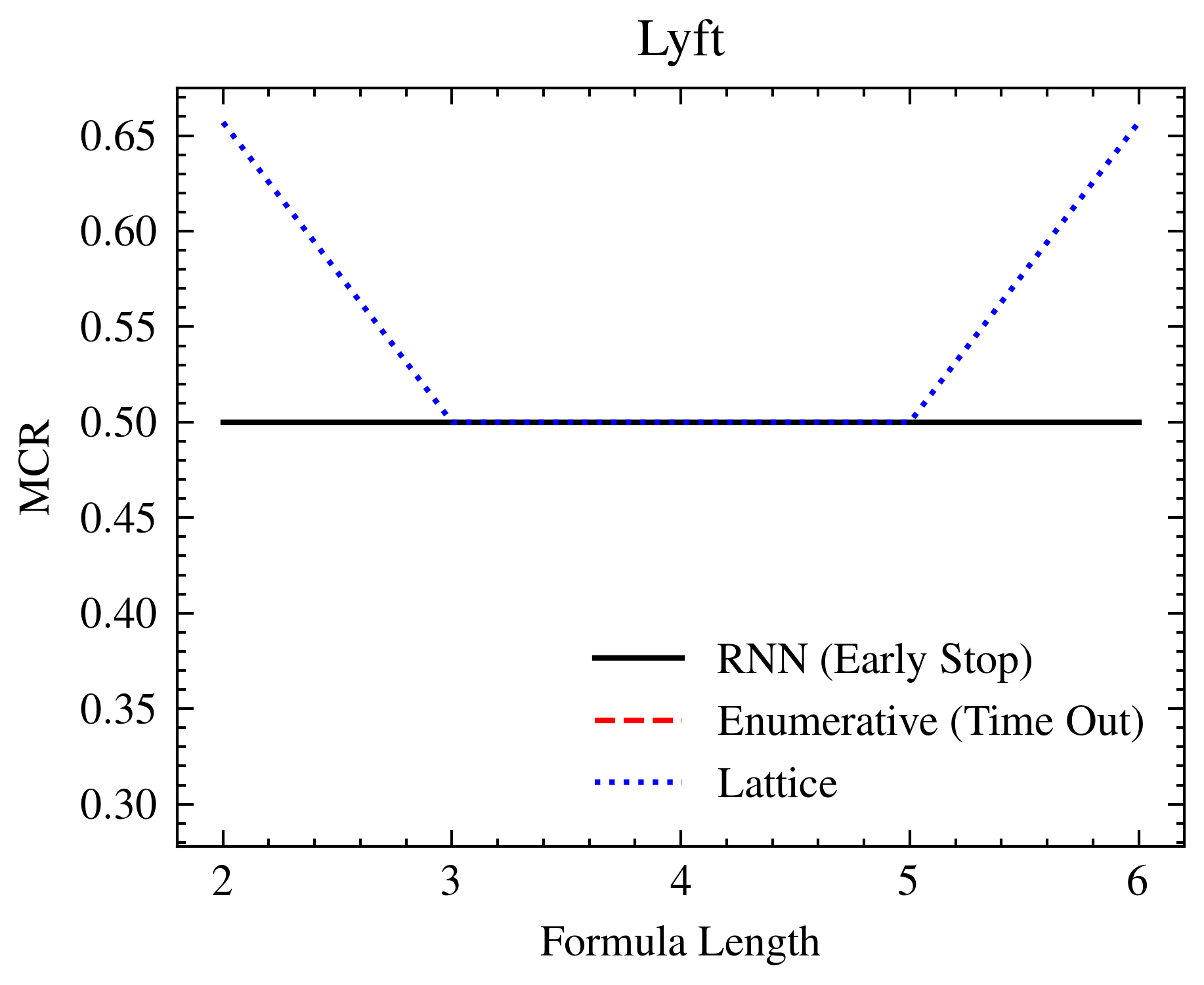}} 
        \\
        {\includegraphics[width=.24\linewidth]{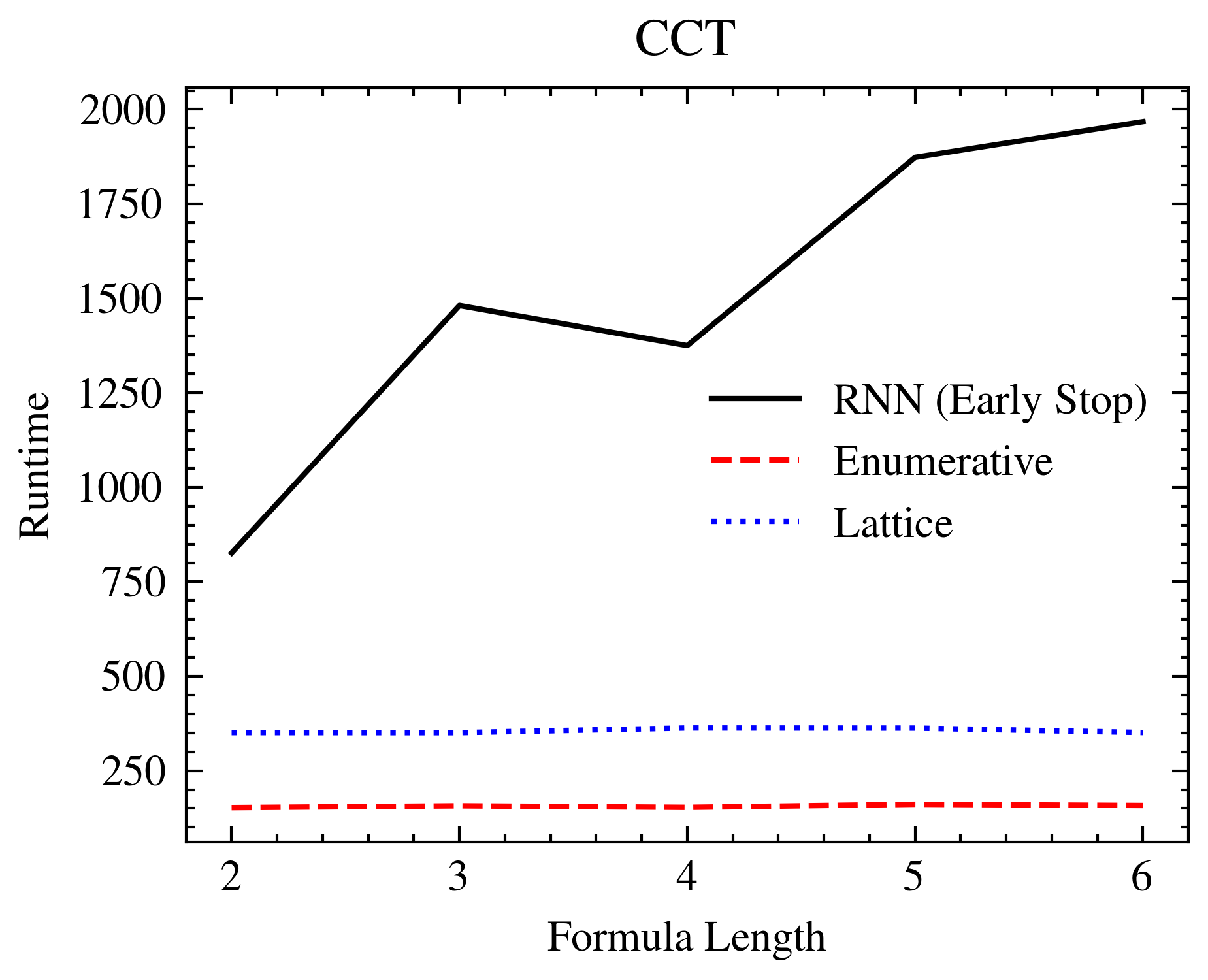}}
        & 
        {\includegraphics[width=.24\linewidth]{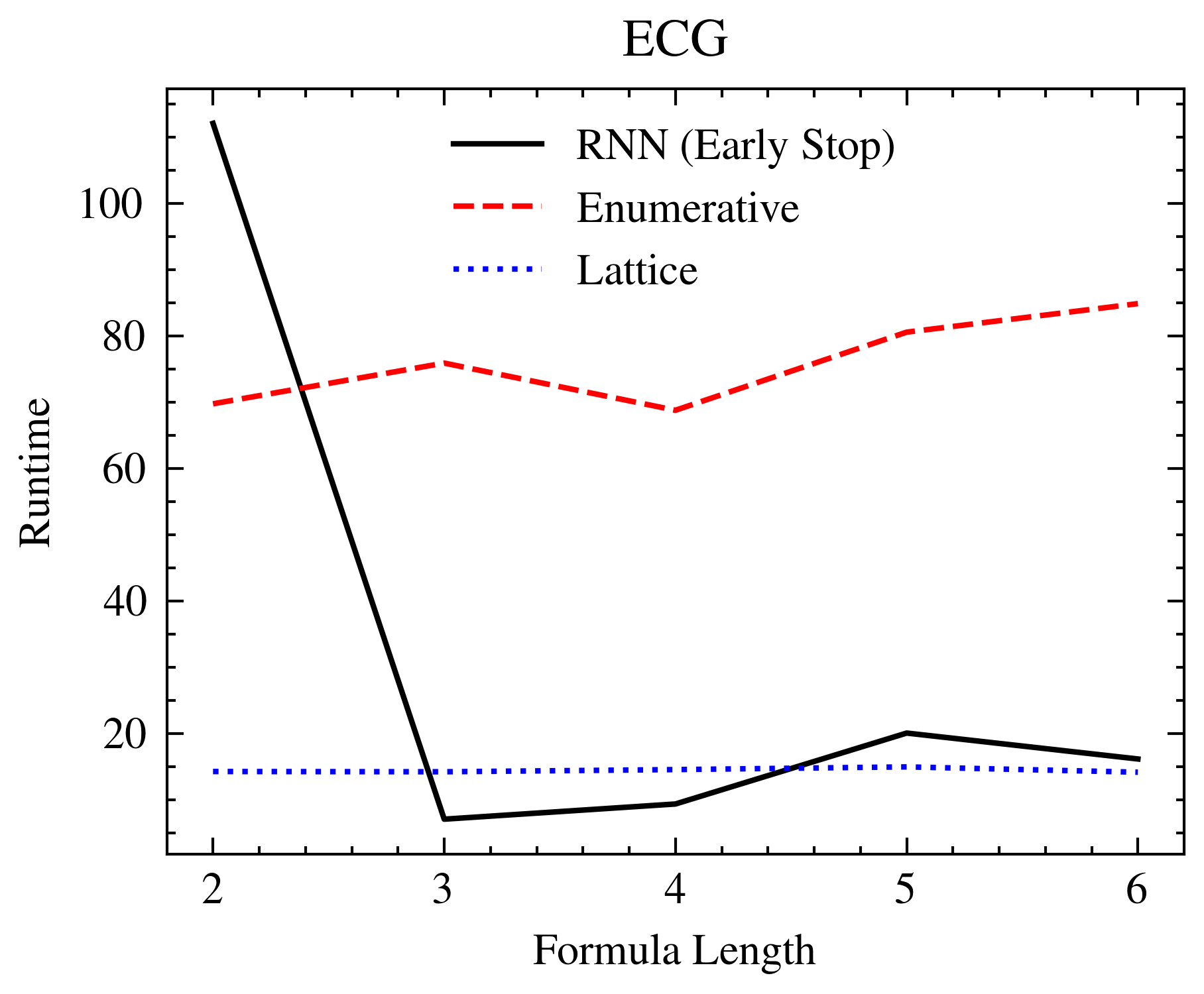}}
        &
        {\includegraphics[width=.24\linewidth]{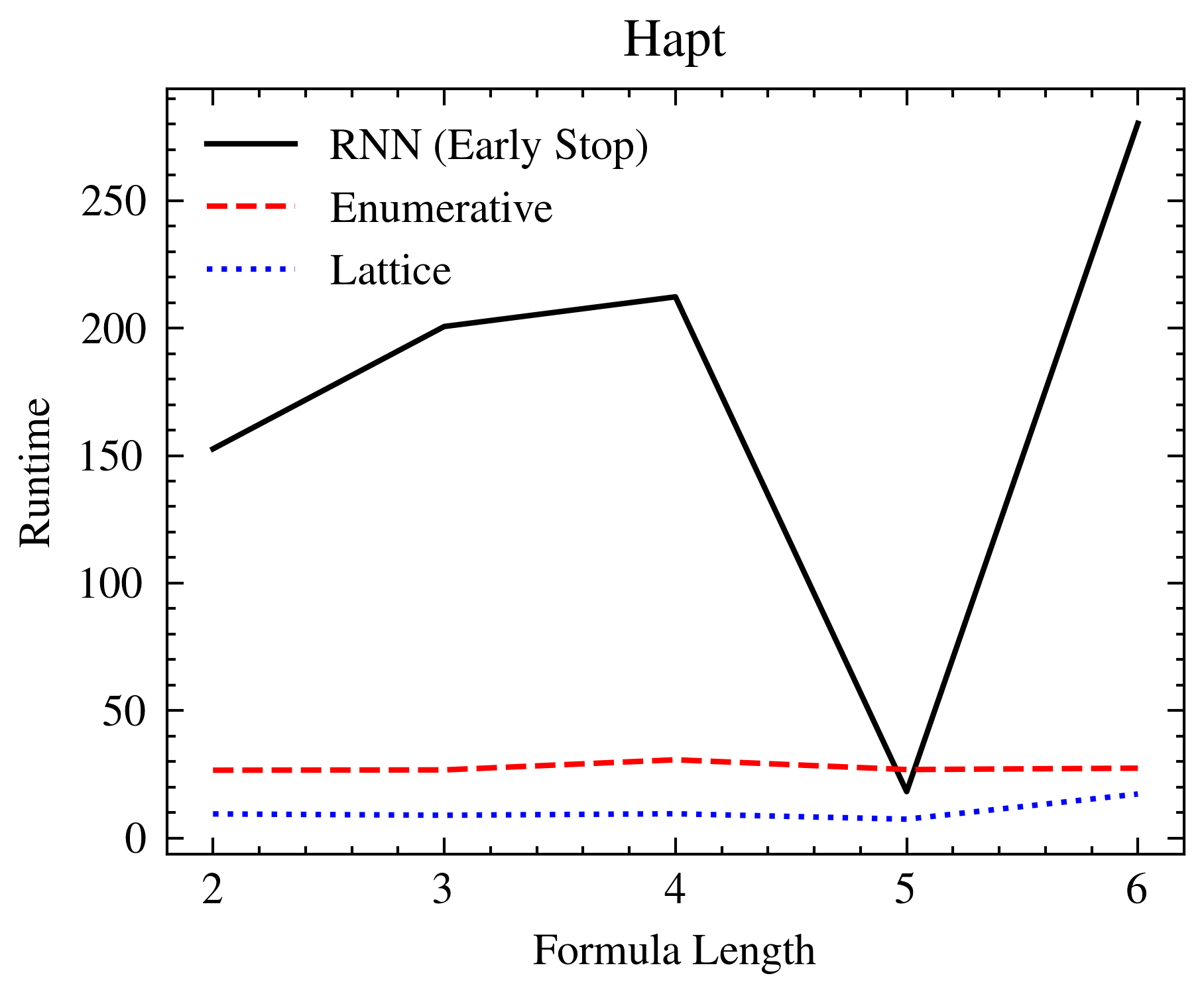}}
        & 
        {\includegraphics[width=.24\linewidth]{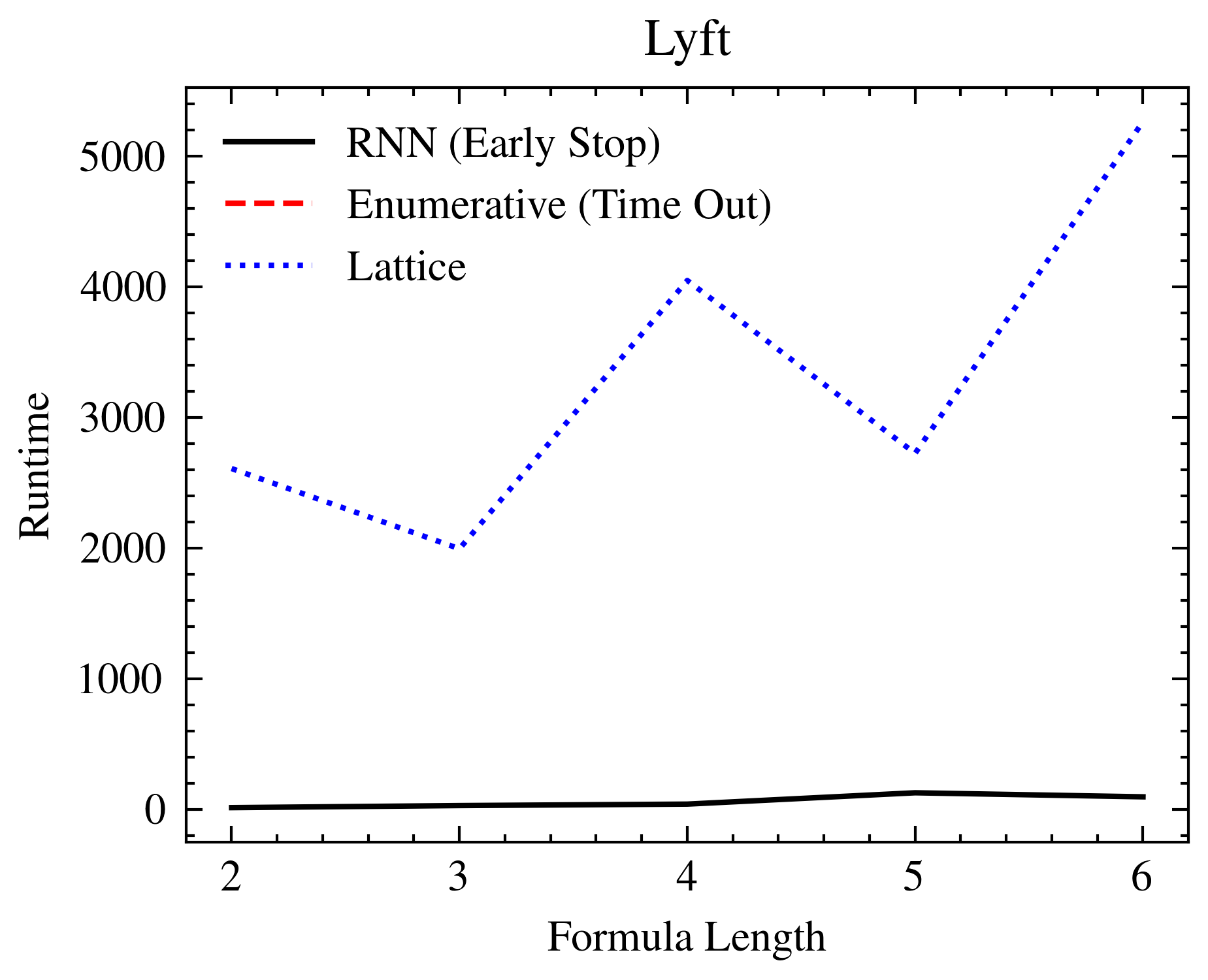}}
    \end{tabular}
    \caption{The MCR (top) and runtime (bottom) of each method on all four datasets. \ourmethod~with early stopping conditions yielded better MCR than the Lattice method and comparable MCR to the Enumerative method on all but the ECG dataset. \edit{Runtime for \ourmethod~was typically worse.  However, for larger datasets such ECG and Lyft, runtime for \ourmethod~was better and still produced results while others timed out.}}
    \label{fig:mcr comparison}
\end{figure*}

\fin{In general, we found that early stopping had little impact on the MCR of \ourmethod, as shown in Table \ref{fig:early stopping comparison}. Figure \ref{fig:mcr comparison} shows the MCR and runtime comparisons for Enumerative, Lattice, and \ourmethod~under early stopping, over all datasets}. \ourmethod~under early stopping often yielded lower MCR than the Lattice method, and \edit{as good or better} MCR than the Enumerative method.  \edit{One exception is the ECG dataset for which \ourmethod~formulas of length $>2$ had much greater MCR than those of the Enumerative method.} For the HAPT and CCT datasets, the early stopping conditions proved to be too loose, allowing \ourmethod~to run substantially longer than the other two methods. For ECG and Lyft, the other methods had longer runtimes than \ourmethod, with the Enumerative method timing out at formula lengths $>2$.

\begin{table}[]
\centering
\begin{tabular}{ll | lll | lll}
                    Dataset  &                & 
                      & {MCR} &       & 
                      & {Runtime} &  \\
    \& Length      &  & No ES   & ES & $\Delta$ & No ES      & ES   & $\Delta$ \\ \hline
ECG  & 2              & 0.10 & 0.09  & 0.20               & 177.3  & 112.1  & 0.58               \\
                      & 3              & 0.47 & 0.47  & 0.00               & 469.5  & 7.0    & 65.7              \\
                      & 4              & 0.47 & 0.47  & 0.00               & 611.9  & 9.3    & 64.6              \\
                      & 5              & 0.47 & 0.50  & -0.07              & 973.2  & 20.0   & 47.6              \\
                      & 6              & 0.47 & 0.50  & -0.07              & 763.0  & 16.1   & 46.3              \\ \hline
Hapt & 2              & 0.00 & 0.00  & 0.00               & 154.7  & 152.7  & 0.01               \\
                      & 3              & 0.00 & 0.00  & 0.00               & 202.8  & 200.6  & 0.01               \\
                      & 4              & 0.00 & 0.00  & 0.00               & 249.9  & 212.3  & 0.2               \\
                      & 5              & 0.00 & 0.10  & -1.00              & 470.3  & 18.2   & 24.8              \\
                      & 6              & 0.00 & 0.00  & 0.00               & 332.3  & 280.3  & 0.2               \\ \hline
CCT  & 2              & 0.03  & 0.03   & 0.00               & 843.1  & 826.2  & 0.02               \\
                      & 3              & 0.03  & 0.03   & 0.00               & 1655.3 & 1481.1 & 0.1               \\
                      & 4              & 0.03  & 0.03   & 0.00               & 1886.9 & 1375.0 & 0.4               \\
                      & 5              & 0.03  & 0.03   & 0.00               & 2879.5 & 1873.1 & 0.5               \\
                      & 6              & 0.03  & 0.03   & 0.00               & 2624.9 & 1967.6 & 0.33               \\ \hline
Lyft & 2              & 0.30 & 0.5    & -0.40              & 1063.8 & 14.2   & 74.0              \\
                      & 3              & 0.50 & 0.50    & 0.00               & 2514.5 & 30.05   & 82.7              \\
                      & 4              & 0.50 & 0.50    & 0.00               & 3510.5 & 41.2   & 84.2              \\
                      & 5              & 0.50 & 0.50    & 0.00               & 5985.1 & 127.9  & 45.8              \\
                      & 6              & 0.50 & 0.50    & 0.00               & 4746.9 & 96.9   & 48.0             
\end{tabular}
\caption{\edit{The MCR and runtime of \ourmethod~with early stopping (ES) and without early stopping (No ES) conditions for formulas, of length 2-6.  Additionally, column $\Delta$ shows the relative improvement of No ES over ES.} In most cases, MCR was similar between the two training schedules. A decrease in MCR is seen for Lyft at length 2 and for HAPT at length 5 when removing early stopping. Improvements in runtime with early stopping were greatest for ECG and Lyft.}
\label{fig:early stopping comparison}
\end{table}

\subsection{Qualitative Comparison of Learned Formulas}
\label{sec: qual comparison}
In this section we take a qualitative look at the best formulas learned by each method for each of the datasets, \edit{including informal inspection of bounds on signal variables.  Because signals are multi-dimensional, we expect the methods to find bounds on different variables.}

\begin{figure}[ht]
	\centering
	\includegraphics[width=\linewidth]{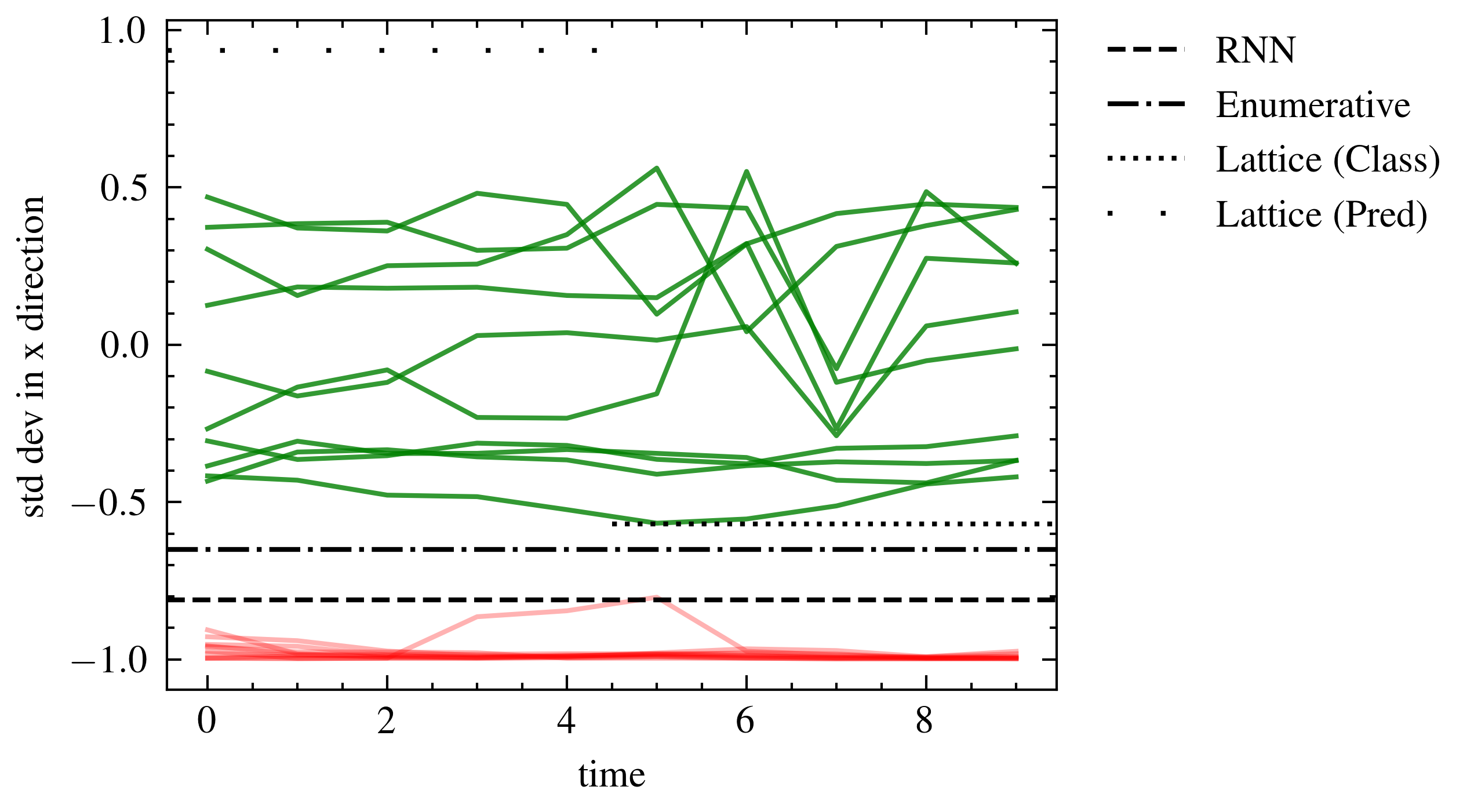}
\caption{Samples of dynamic (green) and static (red) traces of standard deviation of acceleration in the x direction from the HAPT dataset. The decision boundaries of the formulas found by the \ourmethod, Enumerative, and Lattice methods are shown as differently-styled horizontal lines. The prediction and classification sub-formulas found by the Lattice method are plotted separately. Except for the Lattice prediction sub-formula, all boundaries perfectly classify the two sets of traces.}
\label{fig:hapt_results}
\end{figure}

\paragraph{HAPT} There is a clear distinction between the dynamic and static traces in the dataset. Figure \ref{fig:hapt_results} shows a sample of both types of traces alongside the decision boundaries described by the following formulas:

\begin{itemize}
    \item \ourmethod: $\textbf{G} ~ \lnot (x \leq -0.81)$
    \item Enumerative: $\textbf{G} (x > -0.65)$
    \item Lattice: $\textbf{G}_{[1,5)} (x > 0.94) \implies \textbf{F}_{[5,10)} (x > -0.57)$
\end{itemize}

All methods learned that dynamic traces have a signal value greater than some constant that is higher than the signal value of static traces. Both \ourmethod~and Enumerative formulas yield 0.0 MCR.  \edit{As shown in Figure \ref{fig:hapt_results}, \ourmethod~finds a tighter upper bound for the static traces.} For Lattice, only the classification part of formula is able to distinguish between the two sets of traces.

Because the distance between the two classes of traces is large, \ourmethod~found a distinct formula for each formula length from 2 to 6, which all yielded 0.0 MCR. We share the length-2 formula above to compare with the formulas found by the Enumerative and Lattice methods. 

\begin{figure}[ht]
	\centering
	\includegraphics[width=\linewidth]{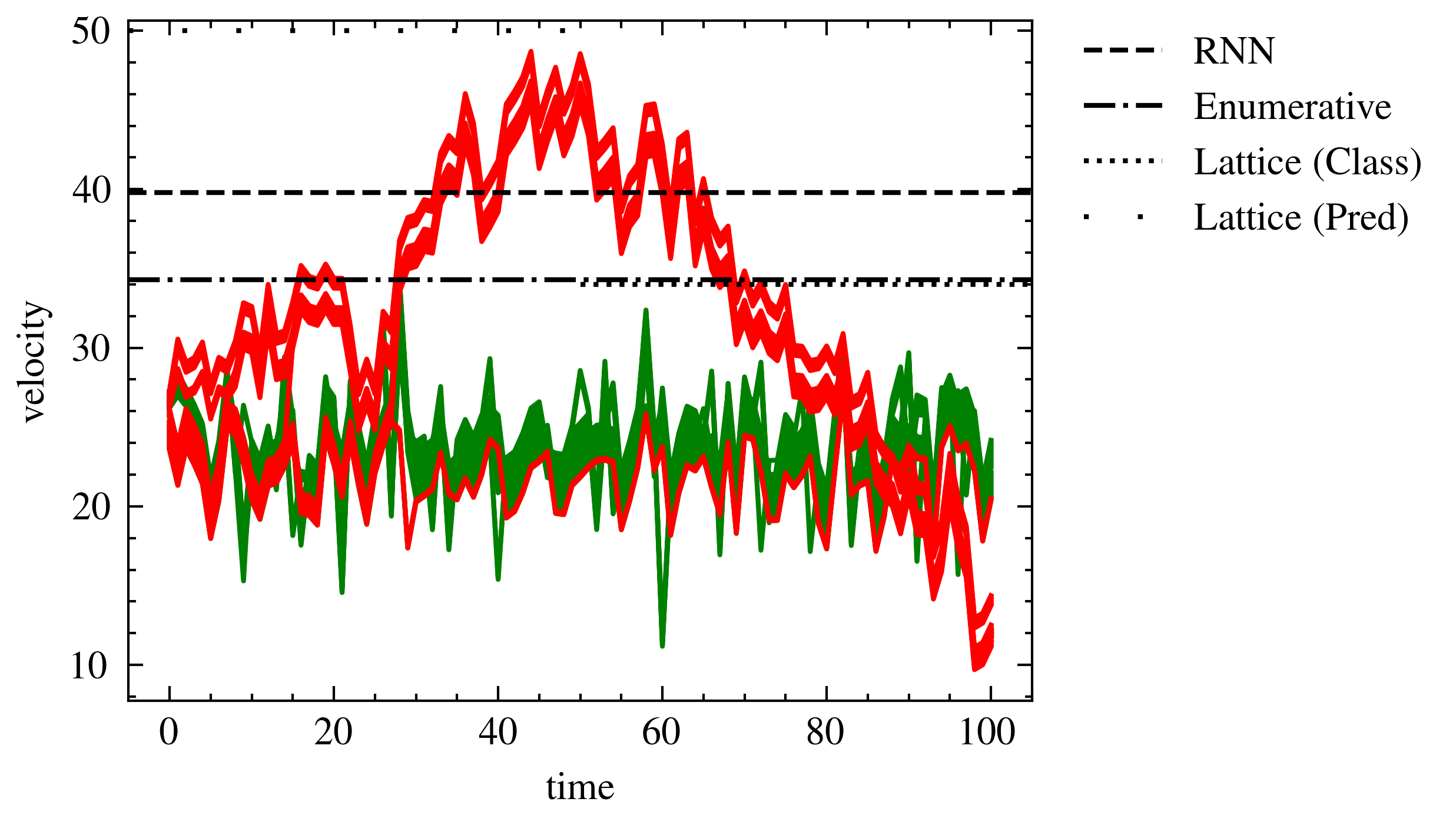}
\caption{Velocity traces from the CCT simulation under normal conditions (green) and anomalous conditions (red). The formulas found by \ourmethod, the Enumerative method, and Lattice Classification method provide reasonable upper bound to distinguish normal traces (between 34-40m/s).}
\label{fig:cruise_control_results}
\end{figure}

\paragraph{CCT} The simulated cruise control train data also shows a clear distinction between the normal and anomalous traces, though there is some overlap. Figure \ref{fig:cruise_control_results} shows the decision boundaries of each of the methods' best formulas alongside the traces.
The formulas are: 
\begin{itemize}
    \item \ourmethod: $\textbf{G} ~ \lnot (v \geq 39.8)$
    \item Enumerative: $\textbf{G} (v < 34.3)$
    \item Lattice: $\textbf{G}_{[1,50.5)} ~ (v > 24.7) \implies \textbf{G}_{[50.5,101)} ~ (v < 34.7)$
\end{itemize}

As with the HAPT data, \ourmethod~found multiple formulas that yield similar MCR of 0.03. Ignoring the prediction part of the Lattice formula, all formulas found that normal behavior for the train meant velocity maintained below a certain value between 34-40m/s. Though \ourmethod~found a looser bound, it yielded the same MCR as the Enumerative formula.

\begin{figure}[t]
	\centering
     \begin{tabular}{c}
        {\includegraphics[width=\linewidth]{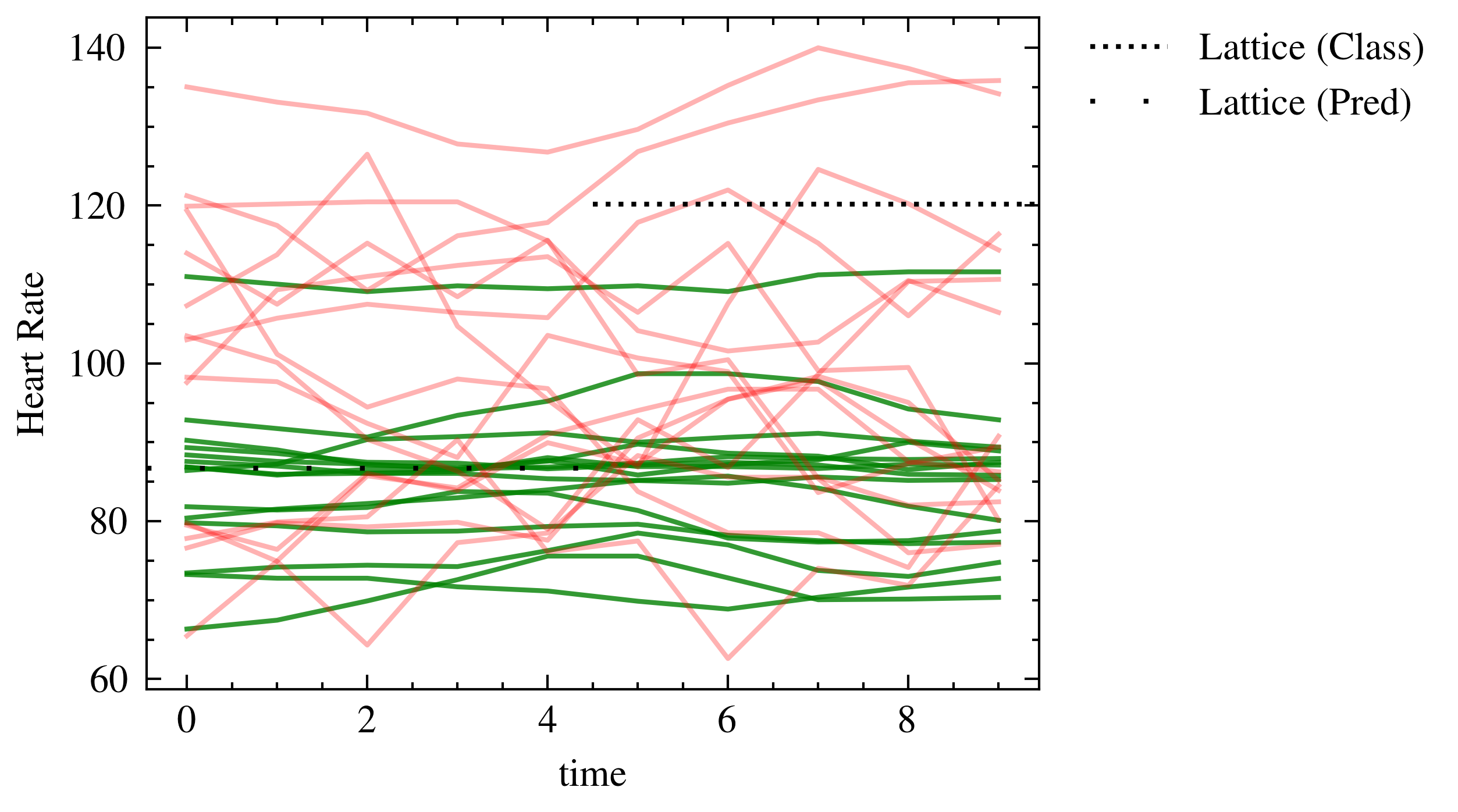}} \\
        {\includegraphics[width=\linewidth]{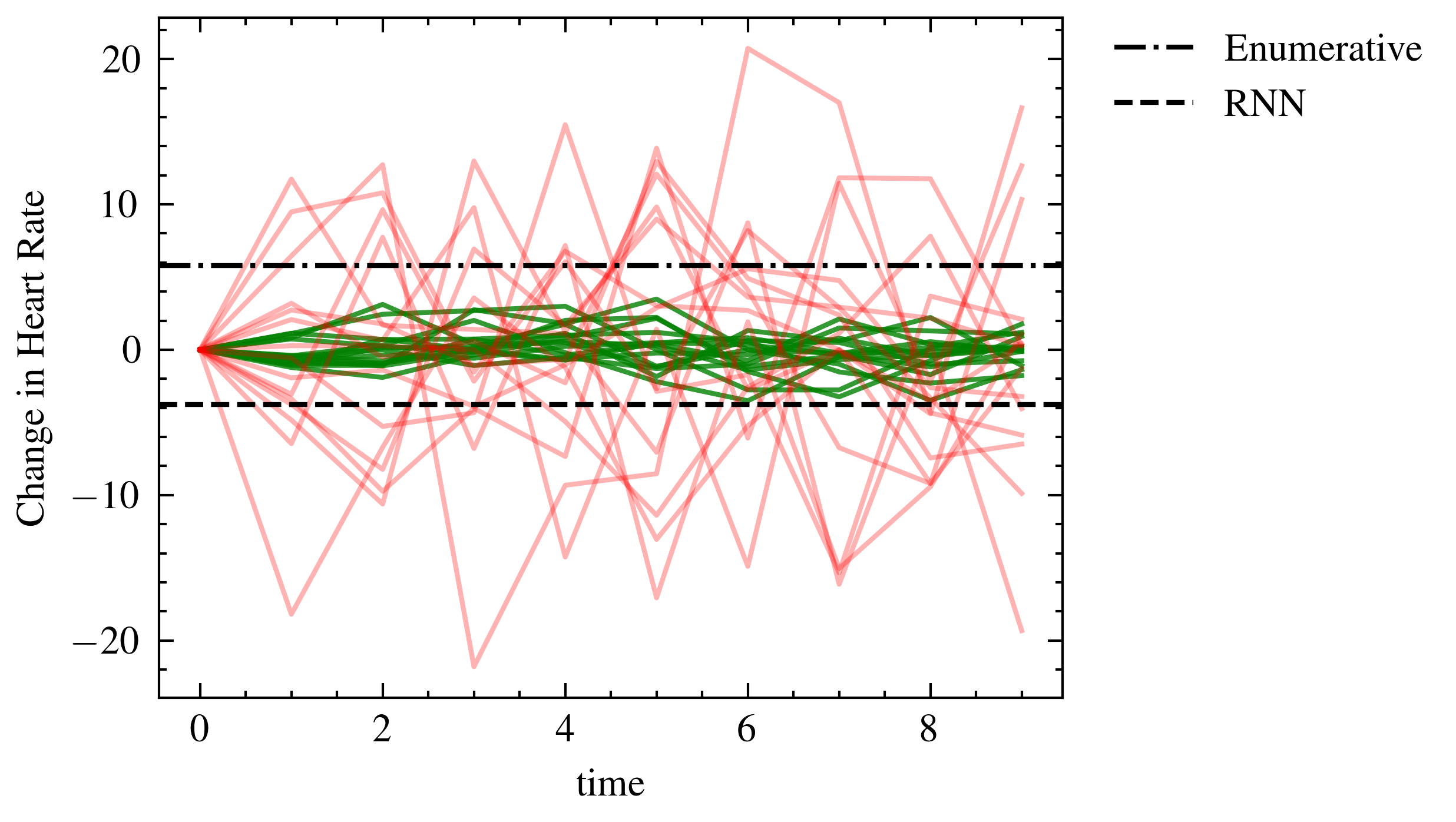}}
    \end{tabular}
\caption{Sample traces of heart rate $hr$ (top) and change in heart rate per second $hr_\textit{diff}$ (bottom) for normal sinus rhythm (green) and atrial fibrillation (red). The Lattice method found bounds for $hr$ while the Enumerative and \ourmethod~methods found bounds for $hr_\textit{diff}$, yielding lower MCR.}
\label{fig:heart_rate_results}
\end{figure}

\paragraph{ECG} Distinguishing a normal sinus rhythm from atrial fibrillation requires looking at both the instantaneous heart range signal $hr$ and the change in heart rate signal $hr_\textit{diff}$.
Figure \ref{fig:heart_rate_results} shows sample traces for both $hr$ and $hr_\textit{diff}$ signals with boundaries found by the learned formulas: 
\begin{itemize}
    \item \ourmethod: $\textbf{G} ~ \lnot (hr_\textit{diff} \leq -3.78)$
    \item Enumerative: $\textbf{G} (hr_\textit{diff} < 5.80)$
    \item Lattice: $\textbf{G}_{[1,5)} ~ (hr > 120.21) \implies \textbf{G}_{[6,10)} ~ (hr < 86.68)$
\end{itemize}

The \ourmethod~and Enumerative methods generate a formula with an atomic proposition using the $hr_\textit{diff}$ signal. \ourmethod~generated a lower bound while the Enumerative algorithm generated an upper bound.  \edit{As shown in Figure \ref{fig:heart_rate_results} the bound found by \ourmethod~is tighter.}  The two formulas yield similar MCR. 
The Lattice method chooses an upper bound for the $hr$ signal in its classification formula. While an upper bound for a normal heart rate is useful, this does not fully \edit{describe} whether the heart is also free of atrial fibrillation. As a result, the Lattice formula yields a much larger MCR.

We also found that longer formulas generated by \ourmethod~for this dataset yielded poor MCR because either the bounds chosen for $hr_\textit{diff}$ were too loose or it got stuck in a local minimum searching for bounds for $hr$. For example, the length 6 formula was $\lnot (\textbf{F} ~ hr \leq 571.061 ~ \land ~ \textbf{F} \textbf{F} ~ hr \geq -25.751)$. This essentially says that $hr$ must be between -25 bpm and 571 bpm, which is not a useful result.

\paragraph{Lyft} By far the most difficult task for all three methods was distinguishing between the true Lyft trajectories and the artificial trajectories. \edit{Under early stopping conditions, \ourmethod~did not yield an MCR better than $0.50$.  However, the best formula found by \ourmethod~under the normal training conditions for length 2 yielded an MCR of $0.302$}. The Enumerative method found a similar formula with slightly better MCR of $0.296$, \edit{but timed out for lengths longer than 2}. The best formula found by the Lattice method yielded MCR of $0.50$. These formulas are:
\begin{itemize}
    \item \ourmethod: $\textbf{F} (y \geq -1.4)$
    \item Enumerative: $\textbf{F} (y < 1.5)$
    \item Lattice: $\textbf{G}_{[1,25)} ~ (x > 50) \implies \textbf{F}_{[26,50)} ~ (x < -10.34)$
\end{itemize}

\begin{figure}[t]
	\centering
     \begin{tabular}{c}
        {\includegraphics[width=\linewidth]{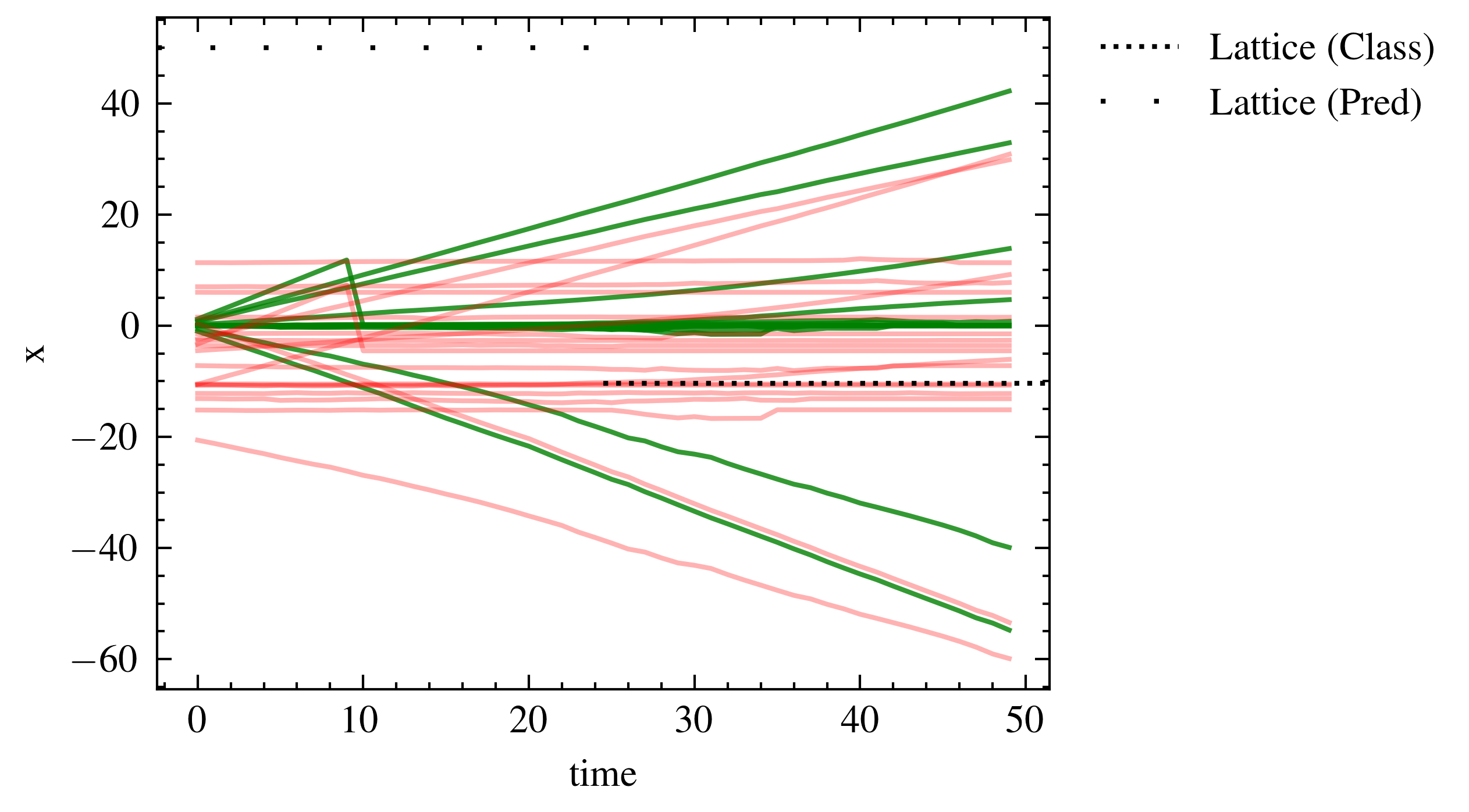}} \\
        {\includegraphics[width=\linewidth]{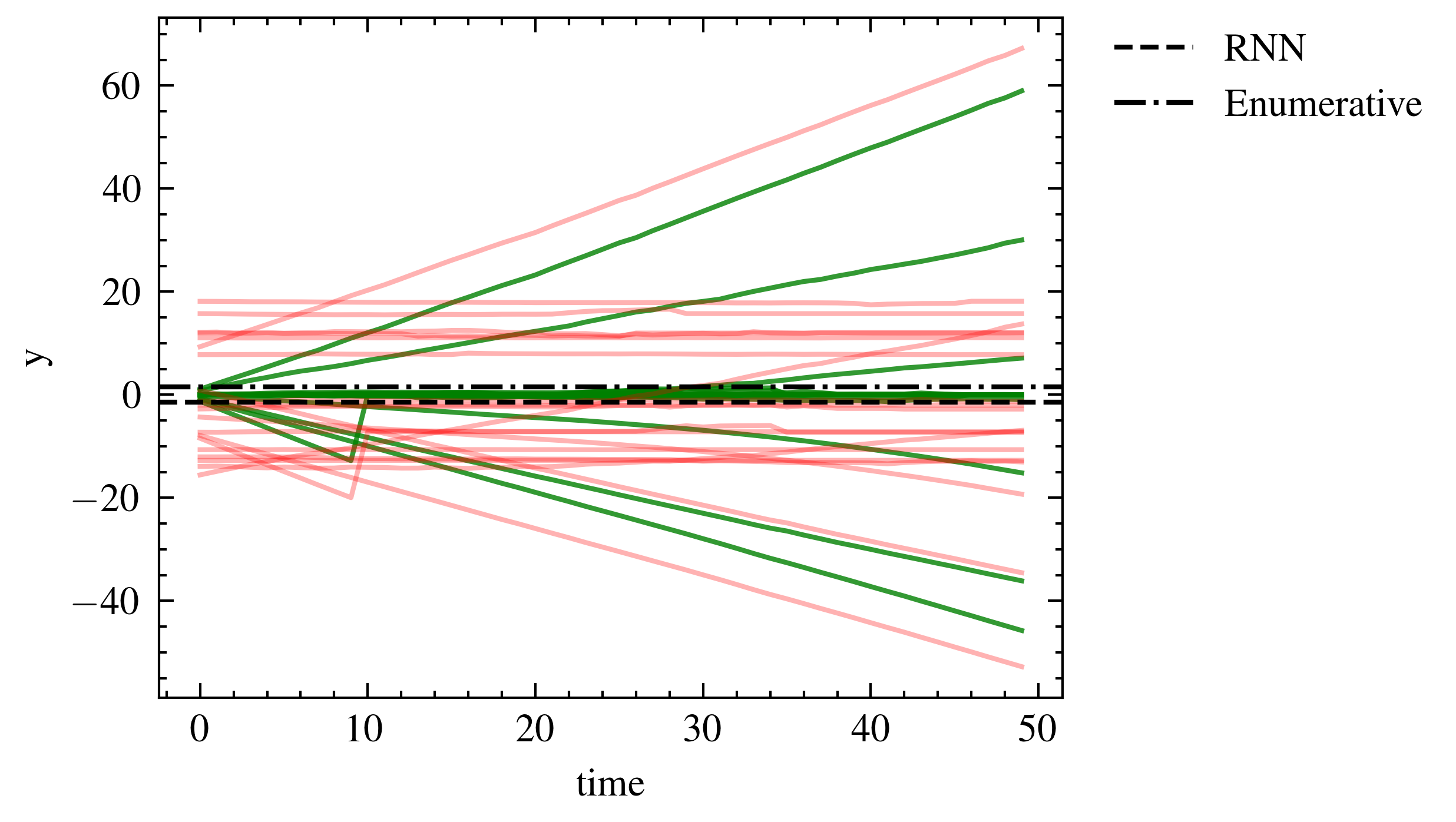}}
    \end{tabular}
\caption{Sample of the traces for $x$ (top) and $y$ (bottom) positions of the Lyft trajectories (green) and artificial trajectories (red). The Lattice method found bounds for $x$ while the Enumerative and RNN methods yielded bounds for $y$. MCR from all methods on this dataset was poor.}
\label{fig:lyft_results}
\end{figure}

Figure \ref{fig:lyft_results} depicts sample traces for the $x$ and $y$ positions for the Lyft trajectories and artificial trajectories with the bounds described in the above formulas. Both $x$ and $y$ signals are similar in that the true and artificial traces have much overlap. Both the \ourmethod~and Enumerative methods \edit{generate loose bounds for $y$} 
(\ourmethod~finds a lower bound while the Enumerative method finds an upper bound) while the Lattice method \edit{generates bounds} for $x$.

All other \ourmethod~tests on the Lyft dataset yielded MCR of 0.5. While structure of the generated formulas varied with length, as expected, so too did the chosen atomic propositions. This leads us to believe that the given features of the dataset are not enough to distinguish a useful classification boundary.

\subsection{Performance Using Continuous Labels}
\edit{The objective in this experiment is to observe how \ourmethod~performs given noisy data.  We accomplish this by using continuous labels instead of binary labels.  For CCT and ECG datasets, we computed robustness values from formulas obtained by the Enumerative method and used these as labels.}

Table \ref{tab:continuous_formulas} shows the best formulas found by the continuous-trained \ourmethod~model compared to the formulas used to generate robustness labels. 
Because the labeling formulas did not have perfect MCR, the theoretical `best' MCR for a \ourmethod trained on these labels would not be 0, but instead would match the labeling formulas' MCR. For CCT, the continuous-trained \ourmethod~yielded a comparable formula to the labeling formula with similar MCR. In contrast, for the ECG, the continuous-trained \ourmethod~found a very different formula, with only slightly worse MCR.

\begin{table*}[]
\centering
\begin{tabular}{lllll}
Dataset              & Labeling Formula         & \edit{Labeling Formula} MCR & \ourmethod~Formula                                                          & \ourmethod~MCR \\
\hline
ECG                  & $\textbf{G} (hr_\textit{diff} < 5.8)$ & 0.069            & $ \lnot ((\textbf{F} hr \geq 19.7) ~\textbf{U}~ (hr_\textit{diff} \leq -4.8))$ & 0.103   \\
CCT &  $\textbf{G}(v \leq 34.3)$                           & 0.030                   & $\textbf{G} \lnot (v \geq 34.2)$                                     & 0.033 
\end{tabular}
\caption{The best formulas found by the continuous-trained \ourmethod~compared to the labeling formulas. For the CCT dataset, a nearly identical formula to the original was found with nearly the same MCR. The formula found for the ECG dataset was longer and more complex with only slightly worse MCR than the original.}
\label{tab:continuous_formulas}
\end{table*}

\subsection{Performance With and Without the Since Operator}
We observed that neither the Enumerative nor Lattice methods generated formulas using the Until operator. Similarly, the \ourmethod~models did not often choose the Since operator. Thus, as another test, we trained the same \ourmethod~architectures described in Section \ref{sec: exp setup} without the $\Si$ layers to see if reducing complexity of the network improves performance. 
Table \ref{tab: results no since} shows that more often than not, removing the $\Si$ layers reduced runtime, as expected, and MCR either remained unchanged or even improved. 
One particularly impressive improvement was in the case of the length 5 formula learned from the ECG data with continuous labels. 
\edit{The original formula learned by \ourmethod~for the ECG dataset, given in Section \ref{sec: qual comparison}  yielded an MCR of 0.466. This formula did not actually use $\Si$ even though it was a choice in the network.}
The formula learned by the network without $\Si$ layers was $\lnot (\textbf{G} \lnot (hr_\textit{diff} \geq -71.33) \land \textbf{G} (hr_\textit{diff} \leq 5.40))$, and yielded 0.034 MCR given the more reasonable bounds found over $hr_\textit{diff}$.

\begin{table*}[!t]
\centering
\begin{tabular}{lllllllll}
 &
  Length = 3 &
   &
  Length = 4 &
   &
  Length = 5 &
   &
  Length = 6 &
   \\
Dataset &
  MCR &
  Runtime &
  MCR &
  Runtime &
  MCR &
  Runtime &
  MCR &
  Runtime
   \\
   \hline
ECG &
  0.47 (0.00) &
  376.19 (-0.20) &
  0.47 (0.0) &
  414.84 (-0.32) &
  0.47 (0.00) &
  903.59 (-0.07) &
  0.47 (0.00) &
  627.67 (-0.18) \\
ECG (Cont. Labels) &
  0.09 (-0.83) &
  363.56 (-0.20) &
  0.41 (-0.11) &
  404.96 (-0.29) &
  0.03 (-0.93) &
  785.06 (-0.16) &
  0.03 (-0.93) &
  590.79 (-0.22) \\
Hapt &
  0.00 (0.00) &
  226.69 (0.12) &
  0.00 (0.00) &
  237.91 (-0.05) &
  0.00 (0.00) &
  413.03 (-0.12) &
  0.00 (0.00) &
  394.42 (0.19) \\
Lyft &
  0.50 (0.00) &
  1936.01 (-0.23) &
  0.50 (0.00) &
  1704.17 (-0.52) &
  0.50 (0.00) &
  4957.23 (-0.17) &
  0.50 (0.00) &
  3149.60 (-0.34) \\
CCT &
  0.03 (0.00) &
  1554.85 (-0.06) &
  0.50 (15.67) &
  1231.70 (-0.35) &
  0.03 (0.00) &
  2287.07 (-0.21) &
  0.50 (15.67) &
  2187.82 (-0.17) \\
CCT (Cont. Labels) &
  0.04 (-0.80) &
  1674.62 (-0.09) &
  0.04 (0.00) &
  1217.47 (-0.42) &
  0.04 (0.00) &
  2470.13 (-0.24) &
  0.50 (13.29) &
  2179.31 (-0.26)
\end{tabular}
\caption{Results of a \ourmethod~trained without the $\Si$ operator for lengths 2-6. Proportion of change in performance over a \ourmethod~trained with $\Si$ operators are in parentheses (lower is better). In all of the above, the \ourmethod~was trained without early stopping. In most cases, runtime is reduced and MCR remains unchanged or shows slight improvement.}
\label{tab: results no since}
\end{table*}





\subsection{Experiment on Choosing Formula Length} \label{sec: exp formula length}
\edit{Up until now, our experiments used \ourmethod~models designed to learn formulas \textit{exclusively} of a certain length.
In this experiment, we train a \ourmethod~to search over formulas \textit{up to} length 6.  This is accomplished by connecting the outputs of each sub-network -- that each learn a formula of a specific length -- to a single final choice block.  In actuality, these connections of sub-networks already exist in the largest network (length 6), as smaller formulas are used to compose larger formulas, but are not themselves choices for the final formula unless specified in initialization of the network. 
When specified, this gives the network the most flexibility in its search over formula structure.  Both the Enumerative and Lattice methods have this flexibility, but require more manual intervention if a specific formula length is desired.  For \ourmethod, this is done simply by setting an additional parameter.}

We use a similar training schedule as described in Section \ref{sec: exp setup} but start with initial learning rate of $0.001$ and terminate early if after $5$ reductions in learning rate, the loss ceases to decrease. The change to a smaller learning rate accounts for the increased complexity of the network. Table \ref{tab: exp3 results} shows the performance results and the learned formulas. The MCR was comparable to \ourmethod~models learning formulas of length $>2$. In most cases, \ourmethod~chooses the simpler formula. 


\begin{table*}[!t]
\centering
\begin{tabular}{llll}
Dataset              & MCR                           & Runtime & Formula                        \\
\hline
ECG                  & 0.47                          & 186.42  & \edit{$\textbf{F} ~ (hr_\textit{diff} \leq 71.0)$} \\
ECG (Cont. Labels) &
  0.50 &
  208.51 &
  $(hr \geq -102.3) ~\textbf{U}~ (hr \geq -60.0)$ \\
Hapt &
  0.00 &
  470.41 &
  $(\lnot (x \geq -0.4) ~ \textbf{U}~ \lnot (x \leq 1.6)) \lor (\textbf{G} ~x \geq -0.8)$ \\
Lyft                 & 0.50                          & 1280.96 & $\textbf{F} ~ (x_\textit{diff} \leq 150.0)$   \\
CCT & 0.03                          & 3170.68 & $\textbf{G} ~ (v \geq 39.8)$       \\
CCT (Cont. Labels) &
  0.04 &
  601.79 &
  $\textbf{G} ~ (v \geq 34.2)$
\end{tabular}
\caption{Results of an experiment in which \ourmethod~was permitted to choose a formula of \emph{up to} length 6. In most cases, \ourmethod~selects the simpler formula (i.e., shorter length).}
\label{tab: exp3 results}
\end{table*}

\subsection{Experiment with Time Bounds}
\label{sec:exp time bounds}
We now show how our method can be used to learn bounded time intervals for temporal operators in addition to the structure and parameters of the formula. This is accomplished by assigning weights to the inputs of the temporal operators and quantizing these weights using the technique described in Section \ref{sec:choice and learning}. For example, for the bounded operator $\Ev_{[a,b]}$, to learn the values of $a$ and $b$, the input to the $\Ev$ cell will be $r[0],\ldots,r[T]$, where $r$ is the cell's input robustness signal \emph{which has been pre-processed to be non-negative}, and $T$ is a hyper-parameter.
Each of these inputs is weighted, so the output of the \ourmethod~cell for this operator is 
\begin{equation*}
    \max(w_0r[0],w_1r[1],\ldots,w_Tr[T])
\end{equation*}
Each weight is learned and quantized during training to either 1 or a large negative value $-M$.
After training, time steps with weight 1 are part of the interval, and those with weight $-M$ are effectively dropped.
Note that the resulting time interval can have `holes' in it, e.g., if $w_1 = w_3 =1$ and $w_2=-M$.
Ours is the only method that can learn such time constraints.

Similarly, for the bounded operator $\Gl_I$, inputs included in the interval are assigned a weight of 1 and all others are assigned a large positive number $M$. 

To test this method, we created an artificial dataset of a signal $\signal$ in which the negatively labeled traces have range $[0, 0.5)$ throughout the interval $I_{-}=[1,6)$, and the positively labeled traces have range $[0.5, 1]$ throughout the interval $I_{+1}=[1,3)$ and range $[0, 0.5)$ throughout interval $I_{+2}=[3,6)$. Using this dataset, we train a \ourmethod~model of length 2 to find a formula with structure $\Gl_I p$ where the parameters of $p$ and the interval $I$ must be learned. The model produced the formula $\Gl_{[1,3]} x \geq 0.42$ which yielded an MCR of 0.08.

\section{Discussion}
\label{sec:discussion}

Even though \ourmethod~is architected to compute robustness, we are able to train it with boolean $\pm 1$ labels and obtain competitive results.
Recalling that robustness is positive for satisfying signals, and negative for violating signals, we can see that the discrete labels are actually $sign(\rho(\signal,t))$.
Thus this is a case where we want to learn a function from its sign; another perspective is that we are dealing with a regression problem where \edit{only noisy robustness values are available}.

It is possible to justify the strong performance of \ourmethod~by a more direct observation: boolean semantics, which evaluate a formula to $\pm 1$, are a special case of robust semantics. 
Namely, if we define an atom's robustness to be $\pm 1$, and keep the rest of Eqs.\eqref{rob semantics negation}-\eqref{rob semantics since} unchanged, we obtain the boolean semantics.
Thus there is reason to believe that supplying the boolean labels would still yield good results, though of course this required the empirical confirmation we have presented.

\edit{Domain knowledge of the model can aid design of the \ourmethod~architecture.  For instance, choice block inputs can be curated to specify which signal dimensions should be included in atomic propositions, or which temporal operators are appropriate for signal constraints.  Additionally, cells can be constructed to impose any known constraints.  As an example, a \ourmethod~model learning on the ECG dataset could include fixed (un-learnable) cells that constrain the search over heart rate bounds within expected human heart rate range.  This would promote learning new constraints that may be useful.}


\edit{In contrast to existing template based methods, \ourmethod~is able to conduct the template search and parameter search simultaneously using gradient descent.  We have shown this method is comparable to the existing ones in that its generated formulas which are equal in simplicity (length) more often than not yield the same MCR or better.} 
\edit{Another strength of our method we have shown is the possibility for learning future time STL formulas by feeding the signal in reverse chronological order, from $t=T$ to $t=0$.}

\edit{It must be noted, however, that training a \ourmethod~takes considerably longer time.  In some cases early stopping can be applied to to reduce training time without loss in performance. For larger datasets like Lyft, this actually reduced runtime to less than that of the Lattice and Enumerative methods, the latter of which timed out.  Thus, for smaller datasets with few features, these methods can be preferred, assuming a monotone formula.  Otherwise, a \ourmethod~model may be more suitable.}

\edit{The fact that different formulas can yield the same MCR highlights that further research is required to discover more appropriate metrics.  In our qualitative examinations, we looked at tightness of bounds, simplicity of the formula in terms of length, and appropriateness of variables chosen for atoms based on knowledge of the dataset.  Future work may consider these aspects when developing new quantitative metrics.}

Additionally, that several learned formulas were `simple' suggests there is still a need for a richer selection of datasets in temporal logic inference research. Of course, we can create artificial datasets, but the real interest is in seeing what learning approaches work on real problems. 

\section{Conclusion}
\label{sec:conclusion}
We demonstrated \ourmethod, an RNN-based learning algorithm for past-time STL, and provide a systematic comparison of this method against existing STL inference techniques.
\edit{The differentiable nature of \ourmethod~enables learning of a formula's structure and parameters simultaneously using gradient descent}, without requiring enumeration or restriction to a fragment of the logic. 
Future work will consider embedding an equivalence check in the RNN creation phase to avoid searching over equivalent formulas, and a more efficient quantized training procedure.

\bibliography{main}
\bibliographystyle{IEEEtran}

\appendices
\section{Quantization in \ourmethod}
For each of the \ourmethod~models trained as described in Section \ref{sec: exp setup}, we also trained an `unquantized' version of the network. That is, the weights in each choice block were real-valued weights and not the one-hot weights as discussed in Section \ref{sec:choice and learning}. Table \ref{tab: quantization comparison} shows minimal performance loss in the use of quantized weights over unquantized weights in \ourmethod~models for these experiments.

\begin{table*}
\centering
\begin{tabular}{lllllll}
Dataset                  & Method            & Length = 2 & Length = 3 & Length = 4 & Length = 5 & Length = 6 \\
\hline
ECG                      & Quant.   & 0.10       & 0.47       & 0.47       & 0.47       & 0.47       \\
                         & Unquant. & 0.05       & 0.78       & 0.05       & 0.05       & 0.09       \\
ECG                      & Quant.   & 0.50       & 0.50       & 0.10       & 0.47       & 0.47       \\
(Cont. Labels) & Unquant. & 0.10       & 0.10       & 0.03       & 0.05       & 0.03       \\
Hapt                     & Quant.   & 0.00       & 0.00       & 0.00       & 0.00       & 0.00       \\
                         & Unquant. & 0.00       & 0.00       & 0.00       & 0.00       & 0.00       \\
Lyft                     & Quant.   & 0.30       & 0.50       & 0.50       & 0.50       & 0.50       \\
                         & Unquant. & 0.31       & 0.50       & 0.50       & 0.07       & 0.50       \\
CCT     & Quant.   & 0.03       & 0.03       & 0.03       & 0.03       & 0.03       \\
                         & Unquant. & 0.03       & 0.03       & 0.03       & 0.03       & 0.03       \\
CCT (Cont. Labels)     & Quant.   & 0.03       & 0.18       & 0.04       & 0.04       & 0.04       \\
 & Unquant. & 0.03       & 0.18       & 0.04       & 0.04       & 0.03      
\end{tabular}
\caption{The effects of quantization in the \ourmethod~choice blocks on MCR for each dataset and each formula length. In all cases, early stopping was not used and the model trained for the full 5000 epochs. In general, the unquantized \ourmethod~ model performs better than its quantized counterpart. For the CCT and Hapt datasets, the difference in MCR is minimal.}
\label{tab: quantization comparison}
\end{table*}

\def \sa {\mathbf{a}}
\def \sb {\mathbf{b}}
\def \sc {\mathbf{c}}

\section{Derivation of the RNN Cell for Since Operator}
This section walks through how we devised the cell architecture for the semantics of the $\Si$ operator. In the following, we simplify some of the notation by writing $a \sqcap b$ for the minimum of $a$ and $b$, and $a\sqcup b$ for the max of $a$ and $b$. 
We also write $\phi[t]$ instead of $\rho_\phi(\signal,t)$, where the signal $\signal$ is fixed.

We present this derivation for discrete-time signals and logic. Extending it to dense time sequences is possible, and requires treating the time-stamps as inputs to the network, which can then learn how time-stamps affect the robustness. 

\textbf{Minimum of sequences:}
Suppose we have 2 sequences of numbers, $\sa$ and $\sb$ (not necessarily of the same length). The max of their respective mins is:
\[M := \max(\min(\sa),\min(\sb))\]
Let $d$ be a number that is appended to both sequences. Then, recomputing the max of their mins is:
\[M' = \max(\min(\sa,d), \min(\sb,d))\]
which can also be done by reusing the previous max $M$:
\begin{eqnarray}
\label{eq:2 seq}
M' &=& \min(d, M)
\end{eqnarray}

Now, recall the robust semantics of $\phi \Si \psi$: 
\begin{equation}
    r[t] = \max_{t' \in [0, t]}\left(
    \min(\psi[t'], \\
    \inf_{t'' \in (t', t]} \phi[t''] )\right) \nonumber
\end{equation}

We can break this down into functions, $f(t, t')$ and $g(t, t')$:
\begin{eqnarray}
    r[t] &=&  \max_{t' \in [0, t]} f(t,t') \nonumber \\
    f(t,t') &=& \min(\psi[t'], \min_{t'' \in (t', t]}\phi[t''] ) \nonumber \\
    &=& \min(\psi( t'), g(t, t')) \nonumber
\end{eqnarray}

As we are working in discrete time, we can rewrite this as
\[f(t,t') = \min(\psi[t'], \phi[t'], \phi[t'+1], \ldots, \phi[t])\]

We now show how to obtain robustness for $t=0,1,2$.

\begin{eqnarray*}
r[0] &=& \min(\phi[0], \psi[0]) \\
r[1] &=& \max(f(1,0), f(1,1)) \\
&=& \max(\min(\psi[0], \phi[0], \phi[1]), \min(\psi[1], \phi[1])) \\
&=& \phi[1] \sqcap \max(\underbrace{\min(\psi[0], \phi[0])}_{r[0] ~\text{By Eq.\eqref{eq:2 seq}}}, \psi[1])\\
&=& \phi[1] \sqcap \max(r[0], \psi[1]) \\
r[2] &=& \max(f(0,2),f(1,2), f(2,2)) \\
&=& \max(\min(\psi[0], \phi[0], \phi[1], \phi[2]), \\
&&\min(\psi[1], \phi[1], \phi[2]), \min(\psi[2], \phi[2])) \\
&=& \phi[2] \sqcap \max(\min(\psi[0], \phi[0], \phi[1]), \\
&&\qquad \min(\psi[1],\phi[1]), \psi[2]) \\
&=& \phi[]2] \sqcap \max(\psi[2], \\
&& \underbrace{\max(\min(\psi[0], \phi[0], \phi[1]), \min(\psi[1], \phi[1])))}_{r[1] }\\
&=& \phi[2] \sqcap \max(r[1], \psi[2]) 
\end{eqnarray*}

More generally, it holds that
\begin{equation}
    r[t] = \phi[t] \sqcap \max(r[t-1], \psi[t])
\end{equation}

\textbf{Proof of Eq. \eqref{eq:2 seq}}
The proof is by case analysis.
If $d \leq \min(\sa)$ and $d \leq \min(\sb)$ then $M' = \max(d,d) = d = \min(d,M)$.

If $d \geq \min(\sa)$ and $d \geq \min(\sa)$ then $M'=M= \min(d,M)$. 

If $\min(\sa) \leq d \leq \min(\sb)$ then $M = \min(\sb)$ and $M' = \max(\min(\sa), d) = d = \min(d,M)$.

By symmetry, the case where $\min(\sb) \leq d \leq \min(\sa)$ is handled in the same way as this last case.

\vfill

\end{document}